\documentclass[11pt]{article}
\usepackage{definitions}
\usepackage{graphicx,paralist,hyperref}

\begin{document}

\title{Randomized Gradient Boosting Machine}

\author{Haihao Lu\thanks{Booth School of Business, University of Chicago
({email:  haihao.lu@chicagobooth.edu}).}
\and
Rahul Mazumder\thanks{MIT Sloan School of Management, Operations Research Center and MIT Center for Statistics. ({email: rahulmaz@mit.edu}).
}}


\date{}

\maketitle

\begin{abstract}
Gradient Boosting Machine (GBM) introduced by Friedman~\cite{friedman2001greedy} is a powerful supervised learning algorithm that is very widely used in practice---it routinely features as a leading algorithm in machine learning competitions such as Kaggle and the KDDCup. In spite of the usefulness of GBM in practice, our current theoretical understanding of this method is rather limited. In this work, we propose Randomized Gradient Boosting Machine (RGBM) which leads to substantial computational gains compared to GBM, by using a randomization scheme to reduce search in the space of weak-learners. We derive novel computational guarantees for RGBM. 
We also provide a principled guideline towards better step-size selection in RGBM that does not require a line search.  Our proposed framework is inspired by a special variant of coordinate descent that combines the benefits of randomized coordinate descent and greedy coordinate descent; and may be of independent interest as an optimization algorithm. 
As a special case, our results for RGBM lead to superior computational guarantees for GBM. Our computational guarantees depend upon a curious geometric quantity that we call Minimal Cosine Angle, which relates to the density of weak-learners in the prediction space. 
On a series of numerical experiments on real datasets,
we demonstrate the effectiveness of RGBM over GBM in terms of obtaining a model with good training and/or testing data fidelity with a fraction of the computational cost.  

\end{abstract}

\section{Introduction}\label{intro}
Gradient Boosting Machine (GBM) \cite{friedman2001greedy} is a powerful supervised learning algorithm that combines multiple weak-learners into an ensemble with excellent predictive performance. It works very well in several prediction tasks arising in spam filtering, online advertising, fraud detection, anomaly detection, computational physics (e.g., the Higgs Boson discovery), etc; and has routinely featured as a top algorithm in Kaggle competitions and the KDDCup \cite{chen2016xgboost}. GBM can naturally handle heterogeneous datasets (highly correlated data, missing data, categorical data, etc) and leads to interpretable models by building an additive model~\cite{friedman2000additive}. It is also quite easy to use with several publicly available implementations: scikit-learn~\cite{pedregosa2011scikit}, Spark MLLib~\cite{meng2016mllib}, LightGBM~\cite{ke2017lightgbm}, XGBoost~\cite{chen2016xgboost}, TensorFlow Boosted Trees~\cite{ponomareva2017tf}, etc. 

In spite of the usefulness of GBM in practice, there is a considerable gap between its theoretical understanding and its success in practice. The traditional interpretation of GBM is to view it as a form of steepest descent in a certain functional space~\cite{friedman2001greedy}. While this viewpoint serves as a good starting point, the framework lacks rigorous computational guarantees, especially when compared to the growing body of literature in first order convex optimization.
There has been some work on deriving convergence rates of GBM---see for example~\cite{bickel2006some,freund2017new,mukherjee2013rate,telgarsky2012primal}, and our discussion in Section~\ref{sec:review}.
Moreover, there are many heuristics employed by practical implementations of GBM that work well in practice---for example, the constant step-size rule and column sub-sampling mechanism implemented in XGBoost~\cite{chen2016xgboost}. However, a formal explanation of these heuristics seems to be lacking in the current literature. This prevents us from systematically addressing important (tuning) parameter choices that may be informed by answers to questions like: how might one choose an optimal step-size, how many weak-learners should one subsample, etc? Building a framework to help addressing these concerns is a goal of this paper. 
In this work we build a methodological framework for understanding GBM and its randomized variant introduced here: Randomized Gradient Boosting Machine (RGBM), by using tools from convex optimization. Our hope is to narrow the gap between the theory and practice of GBM and its randomized variants.
Below, we revisit the classical GBM framework and then introduce RGBM.


\subsection{Gradient Boosting Machine}
We consider a supervised learning problem~\cite{ESLBook}, with $n$ training examples
$(x_{i},y_{i}), i = 1,\ldots,n $
such that $x_{i} \in \RR^p$ is the feature vector of the $i$-th example and
$y_{i} \in \RR$ is a label (in a classification problem) or a continuous response (in a regression problem). In the classical version of GBM~\cite{friedman2001greedy}, the prediction corresponding to a feature
vector $x$ is given by an additive model of the form:
\begin{equation}\label{eq:add-model}
f(x):=\sum_{m=1}^{M}\beta_{j_m}b(x;\tau_{j_m})
\end{equation}
where, each basis function $b(x;\tau)\in \RR$ (also called a weak-learner) is a simple function of the feature vector indexed
by a parameter $\tau$, and $\beta_{j}$ is the coefficient
of the $j$-th weak-learner. Here, $\beta_{j_m}$ and $\tau_{j_m}$ are chosen in an adaptive fashion to improve the data-fidelity (according to a certain rule) as discussed below. Examples of weak-learners commonly used in practice~\cite{ESLBook} include wavelet functions, support vector machines, tree stumps (i.e, decision trees of depth one) and classification and regression trees (CART)~\cite{breiman2017classification}, etc. We assume here that the set of weak-learners is finite with cardinality $K$---in many of the examples alluded to above, $K$ can be exponentially large, thereby posing computational challenges.

\textcolor{black}{Let $\ell(y,f(x))$ be a measure of data-fidelity at the observation $(y,x)$ for the loss function $\ell$, which is assumed to be differentiable in the second coordinate.
A primary goal of machine learning is to obtain a function $f$ that minimizes
the expected loss $\mathbb{E}_{P}(\ell(y,f(x)))$ 
where the expectation is taken over the unknown distribution of $(y,x)$ (denoted by $P$). \emph{One} way to achieve this goal is to consider the empirical loss and \emph{approximately}
minimize it using an algorithm like GBM\footnote{Approximately minimizing the empirical loss function via GBM is empirically found to lead to models with good generalization properties---see for e.g.,~\cite{zhang2005boosting} for some formal explanation (under simple settings). The focus of this paper is on the algorithmic properties of GBM as opposed to its generalization properties.}.
GBM is an algorithm that finds a good estimate of $f$ by approximately minimizing the empirical loss:}
\begin{equation}
\min_{f}~~~~\sum_{i=1}^{n}\ell(y_{i},f(x_{i}))\ \label{eq:loss}
\end{equation}
where, $\ell(y_{i}, f(x_{i}))$ measures data-fidelity for the $i$-th sample $(y_{i}, x_{i})$.
The original version of GBM~\cite{friedman2001greedy} (presented in Algorithm \ref{al:gbm}) can be viewed as applying a steepest descent algorithm to minimize the loss function~\eqref{eq:loss}.
\textcolor{black}{GBM starts from a null model $f\equiv0$ and at iteration $m$ computes the pseudo-residual $r^m$ i.e, the negative gradient of the loss function wrt the prediction. Note that the $i$-th coordinate of $r^m$ is given by $r^m_{i}=-{\partial \ell(y_{i},f^{m}(x_{i}))}/{\partial f^m(x_{i})}$ for $i=1, \ldots, n$.}
GBM finds the best weak-learner that fits $r^m$ in the least squares sense:
\begin{equation}\label{eq:(2)GBM}
    j_{m} = \argmin_{j\in [K] }~\min_\sigma~ \sum_{i=1}^n (r_i^m-\sigma b(x_i;\tau_j))^{2}\ 
\end{equation}
where, $[K]$ is a shorthand for the set $\{1,\ldots,K\}$. (In case of ties in the ``argmin'' operation in~\eqref{eq:(2)GBM}, we choose the one with the smallest index---this convention is used throughout the paper.)
We then add the $j_{m}$-{\tth} weak-learner into the model by using a line search. As the iterations progress, GBM leads to a sequence of models
$\{f^m\}_{m \in [M]}$ (see Algorithm~\ref{al:gbm}), indexed by $m$ (the number of GBM iterations). \textcolor{black}{Each model $f^m$ corresponds to a certain data-fidelity and a (small) number of basis elements with corresponding coefficient weights~\cite{freund2017new,friedman2001greedy}. Together, they control the out-of-sample (or generalization) performance  of the model.} The usual intention of GBM is to stop early i.e., approximately minimize Problem~\eqref{eq:loss}---with the hope that the corresponding model will lead to good predictive performance~\cite{freund2017new, friedman2001greedy,zhang2005boosting}.

\begin{algorithm}[h]
\caption{Gradient Boosting Machine (GBM)~\cite{friedman2001greedy}}\label{al:gbm}

\begin{algorithmic}
\STATE {\bf Initialization.}  Initialize with $f^{0}(x)=0$.\\
For $m=0,\ldots,M-1$ do:\\


\ \ \ (1) Compute pseudo-residual $r^m=-\left[\frac{\partial \ell(y_{i},f^{m}(x_{i}))}{\partial f^m(x_{i})}\right]_{i=1,\ldots,n}.$

\ \ \ (2) Find the best weak-learner: $j_{m} =\argmin_{j\in [K]}\min_\sigma \sum_{i=1}^n (r_i^m-\sigma b(x_i;\tau_j))^{2}$.

\ \ \ (3) Choose the step-size  $\rho_m$ by line-search: $\rho_{m}=\argmin_{\rho}\sum_{i=1}^{n}\ell(y_{i},f^{m}(x_{i})+\rho b(x_i;\tau_{j_m}))$.

\ \ \ (4) Update the model $f^{m+1}(x)=f^{m}(x)+\rho_{m}b(x;\tau_{j_{m}})$.\\
\medskip

\STATE  {\bf Output.}  $f^{M}(x)$.
\end{algorithmic}
\end{algorithm}\medskip

Note that since we perform a line-search, rescaling the prediction vector $[b(x_i;\tau_j)]_{i \in [n]}$ does not change the output of Algorithm \ref{al:gbm}. Hence, without loss of generality, we assume that the prediction vector is normalized throughout the paper.

\smallskip

\begin{ass}\label{ass:normalization}
The prediction vector corresponding to each weak-learner is normalized---that is, for every $\tau$, we have $\sum_{i=1}^n b(x_i;\tau)^{2}=1.$
\end{ass}

\smallskip

\begin{rem}
\textcolor{black}{Note that  Assumption \ref{ass:normalization} is mainly used to simplify the notations and proofs in the paper (that follow). This assumption does not change the convergence guarantees in Theorem \ref{thm:strong} and Theorem \ref{thm:non-strong}.}
\end{rem}

\subsection{Randomized Gradient Boosting Machine}\label{sec:RGBM}
The most expensive step in GBM involves finding the best weak-learner (step (2) in Algorithm \ref{al:gbm}). For example, when the weak-learners are decision trees of depth $d$, finding the best weak-learner requires one to go over $O(n^{2^d-1}p^{2^d-1})$ possible tree splits---this is computationally intractable for medium scale problems, even when $d=1$.

It seems natural (and practical) to use a randomization scheme to reduce the cost associated with 
step (2) in Algorithm \ref{al:gbm}.
To this end, we propose RGBM (see Algorithm \ref{al:rgbm}), where the basic idea is to use a randomized approximation for step \eqref{eq:(2)GBM}. To be more specific, in each iteration of RGBM, we randomly pick a small subset of weak-learners $J$ by some rule (see Section~\ref{sec:rules-select}) and then choose the best candidate from within $J$:
\begin{equation}\label{eq:(3)RGBM}
    j_{m} =\argmin_{j\in J} ~\min_\sigma ~ \sum_{i=1}^n (r_i^m-\sigma b(x_i;\tau_j))^{2}\ .
\end{equation}
If we set $|J|$ (the size of $J$) to be much smaller than the total number of weak-learners $K$, the cost per iteration in RGBM will be much lower than GBM. 
We note that the implementation of XGBoost utilizes a related heuristic (called column subsampling)~\cite{chen2016xgboost}, which has been shown to work well in practice. However, to our knowledge, we are not aware of any prior work that formally introduces and studies the RGBM algorithm---this is the main focus of our paper.

Note that the randomized selection rule we are advocating in RGBM is {\it different} from that employed in the well-known Stochastic Gradient Boosting framework by Friedman \cite{friedman2002stochastic}, in which Friedman introduced a procedure that randomly selects a subset of the \emph{training} examples to fit a weak-learner at each iteration. In contrast, we randomly choose a subset of weak-learners in RGBM. Indeed, both feature and sample sub-sampling are applied in the context of random forests~\cite{liaw2002classification}, however, we remind the reader that random forests are quite different from GBM.



\begin{algorithm}[h]
\caption{Randomized Gradient Boosting Machine (RGBM)}\label{al:rgbm}

\begin{algorithmic}
\STATE {\bf Initialization.}  Initialize with $f^{0}(x)=0$.\\

For $m=0,\ldots,M-1$ do:\\


\ \ \ (1) Compute pseudo-residual $r^m=-\left[\frac{\partial \ell(y_{i},f^{m}(x_{i}))}{\partial f^m(x_{i})}\right]_{i=1,\ldots,n}.$

\ \ \ (2) Pick a random subset $J$ of weak-learners by \emph{some rule} (i.e., one of Type 0 - Type 3)

\ \ \ (3) Find the best weak-learner in $J$: $j_{m} =\argmin_{j\in J}\min_{\sigma} \sum_{i=1}^n (r^m_i-\sigma b(x_i;\tau_j))^{2}$.

\ \ \ (4) Choose the step-size $\rho_m$ by one of the following rules:

\ \ \ \ \ \ $ \bullet $ line-search: $\rho_{m}=\argmin_{\rho}\sum_{i=1}^{n}\ell(y_{i},f^{m}(x_{i})+\rho b(x_i;\tau_{j_m}))$;

\ \ \ \ \ \ $ \bullet $ constant step-size: $\rho_m=\rho \left(\sum_{i=1}^n r_i^m b(x_i;\tau_{j_m})\right)$, where $\rho$ is a constant specified a priori.

\ \ \ (5) Update the model $f^{m+1}(x)=f^{m}(x)+\rho_{m}b(x;\tau_{j_{m}})$.\\

\smallskip

\STATE  {\bf Output.}  $f^{M}(x)$.

\end{algorithmic}
\end{algorithm}



\subsubsection{Random Selection Rules for Choosing Subset $J$} \label{sec:rules-select}
We present a set of selection rules to choose the random subset $J$ in step (2) of Algorithm \ref{al:rgbm}:


\begin{itemize}
    \item[] {\bf [Type 0]:} {\it(Full Deterministic Selection)} We choose $J$ as the whole set of weak-learners. This is a deterministic selection rule.
    
    \item[] {\bf [Type 1]:} {\it(Random Selection)} We choose 
    uniformly at random $t$ weak-learners from all possible weak-learners without replacement---the collection is denoted by $J$.
    
    \item[] {\bf [Type 2]:} {\it(Random Single Group Selection)} Given a non-overlapping partition of the weak-learners, we pick one group uniformly at random and denote the collection of weak-learners in that group by $J$. 
    
    \item[] {\bf [Type 3]:} {\it(Random Multiple Group Selection)} Given a non-overlapping partition of the weak-learners, we pick $t$ groups uniformly at random and let the collection of weak-learners across these groups be $J$.
\end{itemize}

\medskip

\begin{rem}
RGBM with Type 0 selection rule leads to GBM.
\end{rem}

We present an example to illustrate the different selection rules introduced above.

\noindent {\bf Example.} We consider GBM with decision stumps for a binary classification problem. Recall that a decision stump is a  decision tree~\cite{ESLBook} with unit depth. The parameter $\tau$ of a decision stump contains two items: (i) which feature to split and (ii) what value to split on. More specifically, a weak-learner characterized by $\tau=(g, s)$ for $g\in[p]$ and $s\in \RR$ is given by (up to a sign change)
\begin{equation}\label{weak-learn-stmps}
    b(x;\tau = (g, s)) = \left\{
    \begin{array}{cc}
    1 & \text{ if } x_g\le s, \\
    -1 & \text{ if } x_g> s.
    \end{array}
    \right.
\end{equation}
Notice that for a given feature $x_g$ and $n$ training samples, {it suffices to consider at most $n$ different values} 
for $s$ (and equality holds when the feature values are all distinct). This leads to $K=np$ many tree stumps $\{b(x;\tau)\}_\tau$ indexed by $\tau$. For the Type 0 selection rule, we set $J$ to be the collection of all $np$ tree stumps, in a deterministic fashion. As an example of Type 1 selection rule, $J$ can be a collection of $t$ tree stumps selected randomly without replacement from all of $np$ tree stumps. Let $I_g$ be a group comprising of all tree stumps that split on feature $x_g$---i.e., $I_g=\{(g, s)~|~s\}$ for a feature index $g\in [p]$. Then $\{I_g\}_{g\in [p]}$ defines a partition of all possible tree stumps. Given such a partition, an example of the Type 2 selection rule is: we randomly choose $g\in [p]$ and set $J=I_g$. Instead, one can also pick $t$ (out of $p$) features randomly and choose all $nt$ tree stumps on those $t$ features as the set $J$---this leads to an instance of the Type 3 selection rule. Note that a special case of Type~3 with $t=1$ is the Type 2 selection rule.

\begin{figure}[h]
\centering
\begin{tabular}{l c c}
\rotatebox{90}{\sf{{~~~~~~~~~~~~~~training loss}}}
& \includegraphics[width=0.42\textwidth, 
trim =.5cm 0.6cm 1.1cm 1.3cm, clip = true]{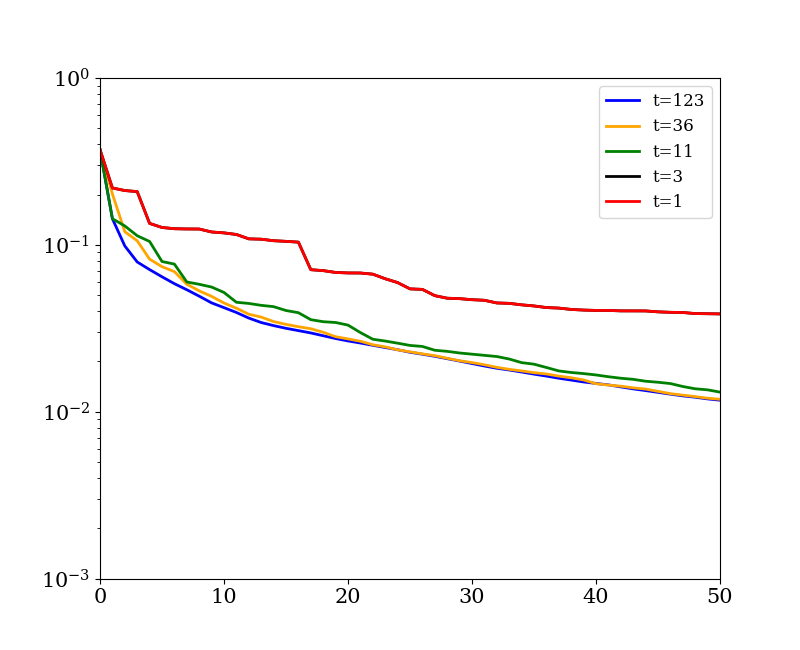} & \includegraphics[width=0.42\textwidth,trim =.5cm 0.6cm 1.1cm 1.3cm, clip = true]{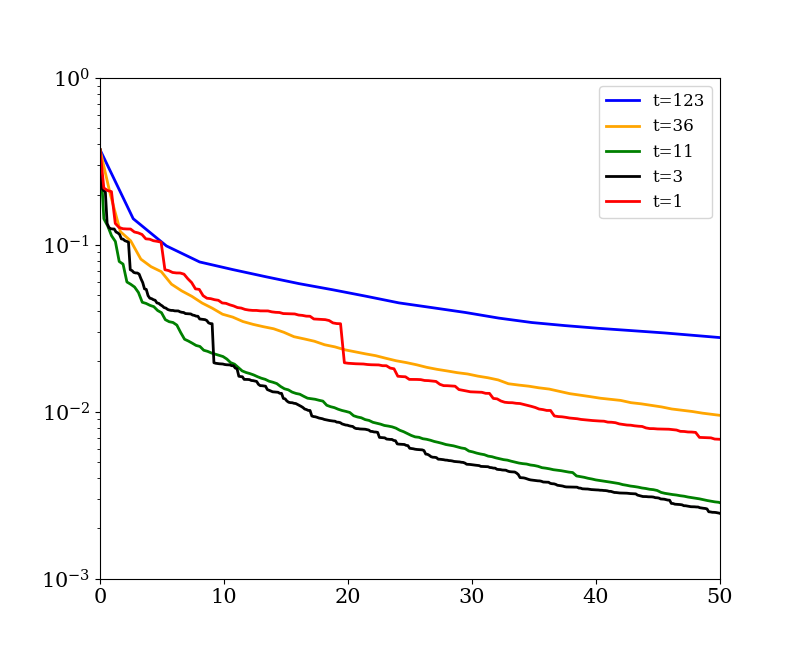} \\
\rotatebox{90}{\sf{{~~~~~~~~~~~~~~testing loss}}}
& \includegraphics[width=0.42\textwidth,
trim =.5cm 0.6cm 1.1cm 1.3cm, clip = true]{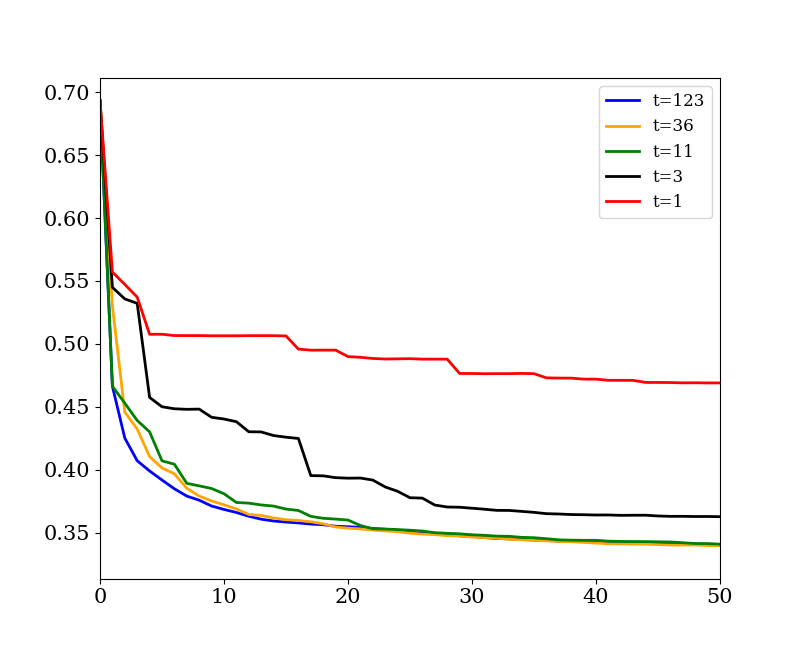} & \includegraphics[width=0.42\textwidth,
trim =.5cm 0.6cm 1.1cm 1.3cm, clip = true]{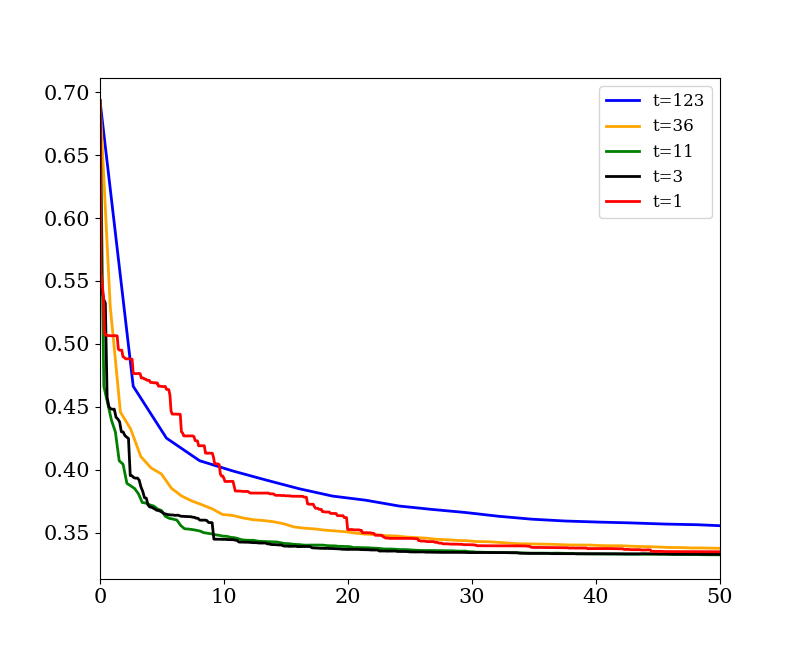} \\
        &  {\sf{{~~~~Iteration}}} & \ \ \ \ \ \  {\sf{{Running Time}}}
\end{tabular}
\caption{Plots showing the training optimality gap in $\log$ scale [top panel] and testing loss [bottom panel] versus number of RGBM iterations and the associated running time (secs) for RGBM with different $t$ values. We consider the a9a dataset (for a classification task) from the LIBSVM library (see text for details). A smaller value of $t$ corresponds to a smaller cost per iteration. As expected we see overall computational savings for a value of $t$ that is smaller than the maximum $t=123$, which corresponds to GBM.}
\label{fig:loss1}
\end{figure}

For motivation, we illustrate the key operating characteristics of RGBM with a real-data example. Figure~\ref{fig:loss1} shows the computational gains of RGBM for solving a binary classification problem with decision stumps. Here we use the Type 3 selection rule (as described above), where each group represents all tree stumps splitting on a single feature, and $G=123$ is the total number of groups. Different lines correspond to different $t$ values---namely, how many groups appear in the random set $J$ in each iteration. The blue line corresponds to GBM (Algorithm \ref{al:gbm}) as it uses all the groups. The major computational cost stems from computing the best weak-learner from a subset of weak-learners. 
The implementation details (leading to Figure~\ref{fig:loss1}) can be found in Section \ref{sec:numerical}. The left column of Figure \ref{fig:loss1} presents the training and testing loss versus number of iterations. 
We can see that when the number of groups $t$ gets smaller, we may get less improvement (in training loss) per iteration, but not by a large margin (for example, the case $t=24$ has a similar behavior as the case $t=123$). The right column of Figure \ref{fig:loss1} shows the training/testing loss versus running time (in seconds). We can see that with a smaller $t$, the cost per iteration decreases dramatically---overall, a small value of $t$ (though, not the smallest) requires less computation (compared to a larger value of $t$) to achieve a similar training/testing error.


\subsection{Related Literature}\label{sec:review}
{\bf Convergence Guarantees for GBM}: 
The development of general convergence guarantees for GBM has seen several milestones in the past decade. After being proposed by Friedman~\cite{friedman2001greedy}, Collins et al~\cite{collins2002logistic} showed the convergence of GBM, without any rates. Bickel and Ritov~\cite{bickel2006some} proved an exponential convergence rate (more precisely $O(\exp(1/\varepsilon^2))$) when the loss function is both smooth and strongly convex. Telgarsky~\cite{telgarsky2012primal} studies the primal-dual structure of GBM. By taking advantage of the dual structure, Telgarsky presents a linear convergence result for GBM with the line search step-size rule. However, the constants in the linear rate are not as transparent as the ones we obtain in this paper, with the only exception being the exponential loss function\footnote{The rate for other loss functions involves a quantity than can be exponentially large in the number of features $p$.}.
There are several works for the convergence rate that apply to specific loss functions. Freund and Schapire \cite{freund1997decision} showed a linear convergence rate for AdaBoost (this can be thought of as GBM with exponential loss and line search rule) under a weak learning assumption. Mukherjee, Rudin and Schapire~\cite{mukherjee2013rate} showed an $O(1/\varepsilon)$ rate for AdaBoost, but the constant depends on the dataset and can be exponentially large in the dimension of the problem. We refer the readers to~\cite{telgarsky2012primal} for a nice review on the early work on Boosting. For LS-Boost (gradient boosting with a least squares loss function), Freund, Grigas and Mazumder~\cite{freund2017new} recently showed a linear rate of convergence, but the rate is not informative when the number of weak-learners is large. Our analysis here provides a much sharper description of the constant---we achieve this by using a different analysis technique.

{\color{black}{Convergence rates of iterative algorithms for classification are closely related to the notion of margins~\cite{freund1996game}. 
Ramdas and Pena~\cite{ramdas2014margins,ramdas2016towards} establish interesting geometric connections between margins, iterative algorithms for classification problems (e.g., the Perceptron and Von-Neumann algorithms)  and condition numbers arising in the study of convex feasibility problems~\cite{cheung2001new,epelman2000condition}. }}

{\color{black}{{\bf Coordinate Descent}: Coordinate descent (CD) methods have a long history in optimization, and convergence of these methods have been extensively studied in the optimization community in the 1980s-90s---see~\cite{bertsekas1989parallel,luo1992convergence,
luo1993error} and~\cite{wright2015coordinate} for a nice overview. There are roughly three types of coordinate descent methods depending on how the coordinate is chosen: randomized, greedy and cyclic CD.
Randomized CD has received considerable attention since the seminal paper of Nesterov~\cite{nesterov2012efficiency}. Randomized CD chooses a coordinate randomly from a certain fixed distribution. \cite{richtarik2014iteration} provides an excellent review of theoretical results for randomized CD. Cyclic CD chooses the coordinates in a cyclic order (see \cite{beck2013convergence} for its first convergence analysis). Recent work shows that cyclic CD may be inferior to randomized CD in the worst case \cite{sun2016worst}---in some examples arising in practice, however, cyclic CD can be better than randomized CD~\cite{beck2013convergence,gurbuzbalaban2017cyclic,hazimeh_mazumder_2019}. In greedy CD, we select the coordinate yielding the largest reduction in the objective function value.  Greedy CD usually delivers better function values at each iteration (in practice), though this comes at the expense of having to compute the full gradient in order to select the coordinate with the largest magnitude of the gradient. On a related note, for the same training data-fidelity (i.e., objective function value), greedy CD usually leads to a model with fewer nonzeros compared to randomized CD---in other words, greedy CD leads to models that are \emph{more} sparse than randomized CD\footnote{We assume here that CD is initialized with a zero solution.}. 

As we will show later, GBM is precisely related to greedy CD. Thus, we focus here on some of the recent developments in greedy CD. \cite{nutini2015coordinate} showed that greedy CD has faster convergence than random CD in theory, and also provided several applications in machine learning where the full gradient can be computed cheaply.  Several parallel variants of greedy CD methods have been proposed in \cite{scherrer2012scaling,scherrer2012feature,you2016asynchronous} and numerical results demonstrate their advantages in practice. \cite{stich2017approximate} 
present an useful scalability idea for steepest CD by maintaining an approximation of the entire gradient vector, which is used to identify the coordinate to be updated. 
More recently, \cite{lu2018accelerating} propose an accelerated greedy coordinate descent method.
}}

\subsection{Contributions}
Our contributions in this paper can be summarized as follows:

~~ 1. We propose RGBM, a new randomized version of GBM, which leads to significant computational gains compared to GBM. This is based on what we call a Random-then-Greedy procedure (i.e., we select a random subset of weak-learners and then find the best candidate among them by using a greedy strategy). In particular, this provides a formal justification of heuristics used in popular GBM implementations like XGBoost, and also suggests improvements. Our framework may provide guidelines for a principled choice of step-size rules in RGBM.

~~ 2. We derive new computational guarantees for RGBM, based on a coordinate descent interpretation. {In particular, this leads to new guarantees for GBM that are superior to existing guarantees for certain loss functions.} The constants in our computational guarantees are in terms of a curious geometric quantity that we call Minimal Cosine Angle---this relates to the density of the weak-learners in the prediction space.

~~ 3. From an optimization viewpoint, our Random-then-Greedy coordinate descent procedure leads to a novel generalization of coordinate descent-like algorithms; and promises to be of independent interest as an optimization algorithm. Our proposal combines the efficiency of randomized coordinate descent and the sparsity of the solution obtained by greedy coordinate descent.

\noindent {\bf Notation:}  For an integer $s$, let $[s]$ denote the set $\{1,2,\ldots,s\}$. For  $a,b\in\RR^p$, $\cos(a,b)$ denotes the cosine of the angle between $a$ and $b$, i.e., $\cos(a,b)={\langle a, b \rangle}/{(\|a\|_2\|b\|_2)}$. Matrix $B$ denotes the prediction for all samples over every possible weak-learner, namely $B_{i,j}=b(x_i;\tau_j)$ for $i\in[n], j\in[K]$. $\Bj$ is the $j$-{\tth} column of $B$ and $B_{i:}$ is the $i$-{\tth} row of $B$. We say $\{I_g\}_{g\in[G]}$ is a partition of $[K]$ if $\cup_{g\in [G]}I_g=[K]$ and $I_g$s are disjoint. We often use the notation $[a_i]_{i}$ to represent a vector $a$.

\section{Random-then-Greedy Coordinate Descent in the Coefficient Space}\label{sec:equiv}

Let $[b(x;\tau_j)]_{j \in [K]}$ be a family of all possible weak-learners. Let
$$
f(x) = \sum_{j=1}^K \beta_j b(x;\tau_{j}) 
$$
be a weighted sum of all $K$ weak-learners $b(x;\tau_j)$, where $\beta_j$ is the coefficient of the $j$-{\tth} weak-learner (we expect a vast majority of the $\beta_{j}$s to be zero). We refer to the space of $\beta\in \RR^K$ as the ``coefficient space''. We can rewrite the minimization problem \eqref{eq:loss} in the coefficient space as:
\begin{equation}\label{eq:loss_beta_1}
\min_{\beta}~~\LC(\beta):=\sum_{i=1}^{n}\ell\left(y_{i},\sum_{j=1}^{K}\beta_{j}b(x_{i};\tau_{j})\right) \ .
\end{equation}
Here, we assume $K$ to be finite (but potentially a very large number). We expect that our results can be extended to deal with an infinite number of weak-learners, but we do not pursue this direction in this paper for simplicity of exposition.

Recall that $B$ is a $n\times K$ matrix of the predictions for all feature vectors over every possible weak-learner, namely $B=\left[b(x_{i};\tau_{j})\right]_{i\in[n],j\in[K]}$. Then each column of $B$ represents the prediction of one weak-learner for the $n$ samples, and each row of $B$ represents the prediction of all weak-learners for a single sample. Thus we can rewrite \eqref{eq:loss_beta_1} as
\begin{equation}\label{eq:loss_over_beta}
    \min_{\beta}\LC(\beta):=\sum_{i=1}^{n}\ell\left(y_{i},\Bi \beta\right) \ .
\end{equation}

Algorithm~\ref{al:rgcd} presents the Random-then-Greedy Coordinate Descent (RtGCD) algorithm for solving~\eqref{eq:loss_over_beta}. We initialize the algorithm with $\beta=0$. At the start of the $m$-{\text{th}} iteration, the algorithm randomly chooses a subset $J$ of the coordinates using one of the four types of selection rules described in Section~\ref{sec:rules-select}. The algorithm then ``greedily'' chooses $j_m \in J$ by finding a coordinate in $\nabla_J\LC(\beta^m)$ with the largest magnitude. We then perform a coordinate descent step on the $j_m$-{\text{th}} coordinate with 
either a line-search step-size rule or a constant step-size rule.

\begin{algorithm}[h]
\caption{Random-then-Greedy Coordinate Descent (RtGCD) in the Coefficient
Space}\label{al:rgcd}

\begin{algorithmic}

\STATE {\bf Initialization.}  Initialize with $\beta^{0}=0$.\\

\smallskip

For $m=0,\ldots,M-1$ do:

~~ {\bf Perform Updates.}  

~~(1) Pick a random subset $J$ of coordinates  by \emph{some rule} (i.e., one of Type 0 - Type 3).

~~(2) Use a greedy rule to find a coordinate in $J$: $j_m=\argmax_{j\in J}|\nabla_{j}\LC(\beta^{m})|$.

~~(3) Choose the step-size  $\rho_m$ by

\ \ \ \ \ \ $ \bullet $ line-search: $\rho_{m}=\argmin_{\rho}\sum_{i=1}^{n}\ell(y_{i},\Bi \beta^m + \rho B_{i,j_m})$;

\ \ \ \ \ \ $ \bullet $ constant step-size: $\rho_m=-\rho\nabla_{j_m}\LC(\beta^m)$ for a given constant $\rho$.

~~(4) Update coefficients: $\beta^{m+1}=\beta^{m}+\rho_{m}e^{j_{m}}$.\\

\smallskip
\smallskip

\noindent {\bf Output.} The coefficient vector $\beta^M$.
\end{algorithmic}
\end{algorithm}

\textcolor{black}{
\begin{rem}
RtGCD forms a bridge between random CD and greedy CD.  
RtGCD leads to greedy CD when $J$ is the set of all coordinates 
and random CD when $J$ is a coordinate chosen uniformly at random from all coordinates.
To our knowledge, RtGCD is a new coordinate descent algorithm; and promises to be of independent interest as an optimization algorithm.  \\
The choice of the group structure (or $J$) depends upon the application. 
For example, in the context of Boosting (using trees as weak learners), the groups are informed by the Boosting procedure---this is usually specified by the practitioner.
In the context of parallel CD algorithms, \cite{scherrer2012feature} propose a method to group coordinates into blocks for algorithmic efficiency---their method updates multiple coordinates within each block. While the context of our work and that of~\cite{scherrer2012feature} are different, it will be interesting to see how ideas in~\cite{scherrer2012feature} can be used with
Algorithm~\ref{al:rgcd} for improved performance. 
\end{rem}}

\smallskip

The following proposition shows that RGBM (Algorithm \ref{al:rgbm}) is equivalent to RtGCD in the coefficient space (Algorithm \ref{al:rgcd}):

\smallskip

\begin{prop}\label{prop:equiv}
Suppose Algorithm~\ref{al:rgbm} makes the same choice of the random set $J$ as Algorithm~\ref{al:rgcd} (in each iteration); and the step-size rules are chosen to be the same in both algorithms. Then the outputs of Algorithm~\ref{al:rgbm} and Algorithm~\ref{al:rgcd} are the same.
\end{prop}

{\bf Proof.}
We will show by induction that $f^m(x)$ in Algorithm \ref{al:rgbm} is the 
same as $\sum_{j=1}^{K}\beta_{j}^{m}b(x;\tau_{j})$ in Algorithm \ref{al:rgcd} for $m=0,1,\ldots,M$. Then Proposition \ref{prop:equiv} holds as a special case for $m=M$.

For $m=0$, we have $f^0(x)=0=\sum_{j=1}^{K}\beta_{j}^{0}b(x;\tau_{j})$. Suppose $f^m(x)=\sum_{j=1}^{K}\beta_{j}^{m}b(x;\tau_{j}),$ then
\begin{equation}\label{eq:equivalence-1}
    \nabla_j \LC(\beta^m)=-\langle \Bj, r^m\rangle\ ,
\end{equation}
where $r^m$ is the pseudo-residual.
In iteration $m$,
the same random subset $J$ is chosen by both algorithms. Next, Algorithm~\ref{al:rgbm} greedily chooses the weak-learner by 
$$j_{m} =\argmin_{j\in J}\min_{\sigma} \sum_{i=1}^n (r^m_i-\sigma b(x_i;\tau_j))^{2}=\argmin_{j\in J}\min_{\sigma}\|r^m- \sigma \Bj\|_2^2 \ .$$ 
Notice that for any $j$, it holds that $\argmin_{\sigma}\|r^m-\sigma \Bj\|_{2}^{2}=\langle \Bj,r^m\rangle$. Hence, we have that
\begin{equation*}
    \begin{array}{lcl}
    j_m     & = & \displaystyle\argmin_{j\in J}~\|r^m- \langle \Bj,r^m\rangle \Bj\|_2^2   =  \argmin_{j\in J} \left(-\frac{1}{2} \langle \Bj,r^m\rangle^2 \right)
         \\ \\
         & = & \displaystyle\argmax_{j\in J}~ |\langle \Bj,r^m\rangle| = \argmax_{j\in J}~ |\nabla_j \LC(\beta^m)| \ ,
        \end{array}
\end{equation*}
where the second equality follows from $\|\Bj\|_2^2=\sum_{i=1}^n b(x_i,\tau_j)^{2}=1$ due to Assumption \ref{ass:normalization} and the last equality utilizes \eqref{eq:equivalence-1}. Therefore, coordinate $j_m$ obtained by Algorithm \ref{al:rgbm} in the $m$-{\tth} iteration is the same as that obtained by Algorithm \ref{al:rgcd}.

Suppose that both algorithms use a step-size based on the line-search rule, then the step-size in Algorithm \ref{al:rgbm} is given by
$$
\rho_{m}=\argmin_{\rho}\sum_{i=1}^{n}\ell(y_{i},f^{m}(x_{i})+\rho b(x_i;\tau_{j_m}))=\argmin_{\rho}\sum_{i=1}^{n}\ell(y_{i},\Bi \beta^m +\rho B_{i,j_m}) \ ,
$$
where we have (by induction hypothesis) that $f^{m}(x_{i})=\Bi \beta^m$. Thus the step-size $\rho_m$ is the same as that chosen by Algorithm \ref{al:rgcd} (with line-search rule). 

Now, suppose both algorithms use a constant step-size rule with the same constant $\rho$. Then the step-size in Algorithm \ref{al:rgbm} is given by:
$$
\rho_m=\rho \left(\sum_{i=1}^n r_i^m b(x_i;\tau_{j_m})\right)=\rho \langle r^m, \Bjm \rangle = -\rho \nabla_{j_m} \LC(\beta^m) \ ,
$$
which is the same step-size as that in Algorithm \ref{al:rgcd} (with constant step-size rule). 

Thus, the step-size $\rho_m$ at the $m$-{\tth} iteration in Algorithm \ref{al:rgbm} is the same as that of Algorithm \ref{al:rgcd} for both step-size rules. Therefore, it holds that
$$
f^{m+1}(x)=f^m(x)+\rho_m b(x;\tau_{j_m}) = \sum_{j=1}^{K}\beta_{j}^{m}b(x;\tau_{j}) + \rho_m b(x;\tau_{j_m}) = \sum_{j=1}^{K}\beta_{j}^{m+1}b(x;\tau_{j}) \ ,
$$
which completes the proof by induction. \qed

\smallskip

\begin{rem}
In the special case when $J$ contains all weak-learners (i.e, with Type 0 random selection rule), Algorithm \ref{al:rgcd} reduces to standard greedy coordinate descent and Proposition \ref{prop:equiv} shows GBM (Algorithm \ref{al:gbm}) is equivalent to greedy coordinate descent in the coefficient space.
\end{rem}

\section{Machinery: Structured Norms and Random Selection Rules}\label{sec:norms}
In this section, we introduce four norms and establish how they relate to the four types of selection rules (for $J$), as described in Section~\ref{sec:rules-select}.

\subsection{Infinity Norm, Ordered $\ell_1$ Norm, $\ell_{1,\infty}$ Group Norm and a Ordered mixed Norm}\label{sec:three-norms}
We introduce the following definitions.

\smallskip

\begin{mydef}
The ``infinity norm'' $\|\cdot\|_{\infty}$ of vector $a\in \RR^K$ is defined as
\begin{equation*}
    \|a\|_{\infty} = \max_{j \in [K]}|a_j|\ .  ~~~~~~~~~~~~~~~(\text{Infinity norm})
\end{equation*}
\end{mydef}

\smallskip

\begin{mydef}
The ``ordered $\ell_1$ norm'' $\|\cdot\|_{\mS}$ with parameter $\gamma\in \RR^K$ of  vector $a\in \RR^K$ is defined as
\begin{equation*}
    \|a\|_{\mS} = \sum_{j=1}^K \gamma_i|a_{(j)}|\ , ~~~~~~~~~~(\text{Ordered $\ell_1$ norm})
\end{equation*}
where the parameter $\gamma$ satisfies $\gamma_1\ge\gamma_2\ge\ldots\ge\gamma_K\ge 0$ with $\sum_{j=1}^K \gamma_j=1$; and $|a_{(1)}|\ge |a_{(2)}|\ge\ldots\ge |a_{(K)}|$ are the decreasing absolute values of the coordinates of $a$.
\end{mydef}

\smallskip

\begin{mydef}
If $\{I_g\}_{g\in [G]}$ is a partition of $[K]$, then the ``$\ell_{1,\infty}$ group norm'' of vector $a\in \RR^K$ is defined as
\begin{equation*}
\|a\|_{\GG} = \frac{1}{G}\sum_{g=1}^G \|a_{I_g}\|_{\infty}, ~~~~~~~~~~~~~~~~ \text{(Group norm)}
\end{equation*}
 where, $\|a_{I_g}\|_{\infty}$ is the infinity-norm of $a_{I_g}$ (i.e., the sub-vector of $a$ restricted to $I_g$) for $g \in [G]$.
\end{mydef}

\smallskip

\begin{mydef}
If $\{I_g\}_{g\in [G]}$ is a partition of $[K]$, then the ``ordered mixed norm''\footnote{The name stems from the fact that it is a combination of the ordered $\ell_1$ norm and the $\ell_{1,\infty}$ group norm.} with parameter $\gamma\in \RR^G$ 
of vector $a\in \RR^K$ is defined as
\begin{equation*}
    \|a\|_{\CC} = \sum_{g=1}^G \gamma_g\|a_{I_{(g)}}\|_{\infty}\ , ~~~~~~~~~~(\text{Ordered mixed norm})
\end{equation*}
where the parameter $\gamma$ satisfies $\gamma_1\ge\gamma_2\ge\ldots\ge\gamma_G\ge 0$ and $\sum_{g=1}^G \gamma_g=1$. Note that $\|a_{I_{(1)}}\|_{\infty}\ge \|a_{I_{(2)}}\|_{\infty}\ge\ldots\ge \|a_{I_{(G)}}\|_{\infty}$
are the sorted values of $\|a_{I_g}\|_{\infty}, g \in [G]$.
\end{mydef}

\smallskip
\smallskip

\begin{rem}
Note that the group norm~\cite{negahban2008joint} and ordered $\ell_1$ norm (arising in the context of the Slope estimator)~\cite{bogdan2015slope} appear as common regularizers in high-dimensional linear models. In this paper however, they arise in a very different context---see Section~\ref{sec:random-then-greedy}. 
\end{rem}

It can be easily seen that the ordered $\ell_1$ norm is a special instance of the ordered mixed norm where each group contains one element, and the $\ell_{1,\infty}$ group norm is another special instance of the ordered mixed norm where the parameter $\gamma_g\equiv {1}/{G}$ for $g\in[G]$.

With some elementary calculations, we can derive the dual norms of each of the above norms.

\smallskip

\begin{prop}\label{prop:dual-norm} 

    (1) The dual norm of the ordered $\ell_1$ norm is
    \begin{equation}\label{eq:dual-slope}
    \|b\|_{\mS^*}=\max\limits_{1\le i\le K} \frac{\sum_{j=1}^{i} |b_{(j)}|}{\sum_{j=1}^{i} \gamma_j}\ .
    \end{equation}
    
    
    (2) The dual norm of the $\ell_{1,\infty}$ group norm is
    \begin{equation*}
        \|b\|_{\GG^*} = G~\max_{1\le g\le G} \|b_{I_g}\|_1 \ .
    \end{equation*}
    
    (3) The dual norm of the ordered mixed norm is
    \begin{equation*}
        \|b\|_{\CC^*} = \max_{1\le g\le G} \frac{\sum_{j=1}^{g}\|b_{I_{(j)}}\|_{1}}{\sum_{j=1}^{g}\gamma_{j}} \ ,
    \end{equation*}
    where $\|b_{I_{(1)}}\|_{1}\ge \|b_{I_{(2)}}\|_{1} \ge \cdots \ge \|b_{I_{(G)}}\|_{1}$ are the values of $\|b_{I_g}\|_1, g \in [G]$, sorted in decreasing order. 

\end{prop}

\smallskip
\smallskip

\begin{rem}
The proof for part (1) of Proposition \ref{prop:dual-norm} can be found 
in Theorem 1 in~\cite{zeng2014decreasing}. The proof of part (2) is straightforward, and the proof for part (3) follows from that of (1) and (2).
\end{rem}

\subsection{Random-then-Greedy Procedure}\label{sec:random-then-greedy}
Here we introduce a Random-then-Greedy (RtG) procedure that uses a randomized scheme to 
deliver an approximate maximum of the absolute entries of a vector $a\in \RR^K$.
The expected value of the (random) output available from the RtG procedure with four types of selection rules (cf Section~\ref{sec:rules-select}) can be shown to be related to the four norms introduced in Section~\ref{sec:three-norms}.

Formally, the RtG procedure is summarized below:

\begin{center}
    \begin{center}
        {\bf Random-then-Greedy (RtG) procedure}
    \end{center}
    \begin{minipage}[t]{10cm}
    Given $a\in \RR^K$,
    
    1. Randomly pick a subset of coordinates $J\subseteq [K]$.
    
    2. Output $\hj = \argmax_{j\in J}|a_j|$ and $|a_{\hj}|$.
    \end{minipage}
\end{center}


We will next obtain the probability distribution of $\hj$, and the expectation of $|a_{\hj}|$.

Let $J$ be chosen by Type 1 selection rule, namely $J$ is given by a collection of $t$ coordinates, chosen uniformly at random from $[K]$ without replacement. A simple observation is that the probability of a coordinate $j$ being chosen depends upon the magnitude of $a_{j}$ \emph{relative} to the other values $|a_{i}|, i \neq j$; and not the precise values of the entries in $a$.
Note also that if the value of $|a_j|$ is higher than others, then the probability of selecting $j$ increases: this is because (i) all coordinate indices in $[K]$ are equally likely to appear in $J$, and (ii) coordinates with a larger value of $|a_j|$ are chosen with higher probability.
The following proposition formalizes the above observations and presents the probability of a coordinate being chosen.


\smallskip

\begin{prop}\label{prop:prob-R1}
Consider the RtG procedure for approximately finding the maximal coordinate of $a\in\RR^K$ (in absolute value). Recall that $(j)$ is the index of the $j$-{\tth} largest coordinate of $a$ in absolute value\footnote{In case of ties, we choose the smallest index.}, namely $|a_{(1)}|\ge |a_{(2)}|\ge\cdots\ge |a_{(K)}|$. If the subset $J$ is chosen by the Type 1 selection rule, the probability that $(j)$ is returned is
\begin{equation}\label{eq:gtKj}
P\left( \hj = {(j)}\right):=\gtKj=\frac{\binom{K-j}{t-1}}{\binom{K}{t}}
\ .
\end{equation}

\end{prop}
{\bf Proof.}
There are $\binom{K}{t}$ different choices for the subset $J$, and each subset is chosen with equal probability. The event ${\hj}={(j)}$ happens if and only if $(j)\in J$ and the remaining $t-1$ coordinates are chosen from the $K-j$ coordinates. 
There are $\binom{K-j}{t-1}$ different choices of choosing such a subset $J$, which completes the proof of Proposition \ref{prop:prob-R1}. \qed

\smallskip

\begin{rem}
Note that $\gtKj$ is monotonically decreasing in $j$ for fixed $K,t$ (because $j\rightarrow \binom{K-j}{t-1}$ is monotonically decreasing in $j$). This corresponds to the intuition that the RtG procedure returns a coordinate $j$ with a larger magnitude of $a_j$, with higher probability.
\end{rem}

For most cases of interest, the dimension $K$ of the input vector is very large. When $K$ is asymptotically large, it is convenient to consider the distribution of the quantile $q=j/K$ (where $0<q<1$), instead of $j$. The probability distribution of this quantile evaluated at $j/K$ is given by $K\gtKj$.
The following proposition states that $K\gtKj$ asymptotically converges to $t(1-q)^{t-1}$, the probability density function of the Beta distribution with shape parameters $(1,t)$ i.e., $\text{Beta}(1,t)$.

\smallskip

\begin{prop} We have the following limit for a fixed $q \in (0,1)$:
\begin{equation*}
    \lim_{j,K\rightarrow \infty,~j/K=q} K\gtKj = t(1-q)^{t-1} \ .
\end{equation*}
\end{prop}
\noindent {\bf Proof.} 
By using the expression of $\gtKj$ and canceling out the factorials, it holds that
\begin{equation*}
    \begin{array}{lcl}
     \gtKj    & = & \frac{\binom{K-j}{t-1}}{\binom{K}{t}} = \frac{t}{K} \frac{(K-t)(K-t-1)\cdots (K-j-t+2)}{(K-1)(K-2)\cdots(K-j+1)} \\ \\
         & = &  \frac{t}{K} \left(1-\frac{t-1}{K-1}\right) \left(1-\frac{t-1}{K-2}\right)\cdots \left(1-\frac{t-1}{K-j+1}\right).
    \end{array}
\end{equation*}
Denote $A^{K}_t(j)=\left(1-\frac{t-1}{K-1}\right) \left(1-\frac{t-1}{K-2}\right)\cdots \left(1-\frac{t-1}{K-j+1}\right)$, then it holds that
\begin{equation*}
    \begin{array}{lcl}
    \displaystyle\lim_{j,K\rightarrow \infty,~ j/K=q} \ln A^{K}_t(j)     & = & \displaystyle\lim_{j,K\rightarrow \infty,~ j/K=q} \sum_{l=1}^{j-1} \ln \left(1-\frac{t-1}{K-l}\right)  \\ \\
         & = & \displaystyle\lim_{j,K\rightarrow \infty,~ j/K=q} \sum_{l=1}^{j-1} -\frac{t-1}{K-l} \\ \\
         & = & \displaystyle\lim_{j,K\rightarrow \infty,~ j/K=q} (t-1) \ln \left(\frac{K-j}{K}\right) \\ \\
         & = & (t-1) \ln (1-q)\ ,
    \end{array}
\end{equation*}
where the second equality uses $\ln\left( 1-\frac{t-1}{K-l}\right) \approx -\frac{t-1}{K-l}$ and the third equality is from $\sum_{l=1}^{j-1} \frac{1}{K-l} \approx \ln K- \ln(K-j) =\ln (\frac{K}{K-j})$, when both $j,K$ are large and $j/K\approx q$.
Therefore,
$$
\lim_{j,K\rightarrow \infty, j/K=q} K\gtKj = t \lim_{j,K\rightarrow \infty, j/K=q} \exp\left(\ln A^{(K,j)}_t\right) = t(1-q)^{t-1} \ ,
$$
which completes the proof. \qed

\smallskip

\begin{figure}
    \centering
    \scalebox{.95}{\includegraphics[width=9cm]{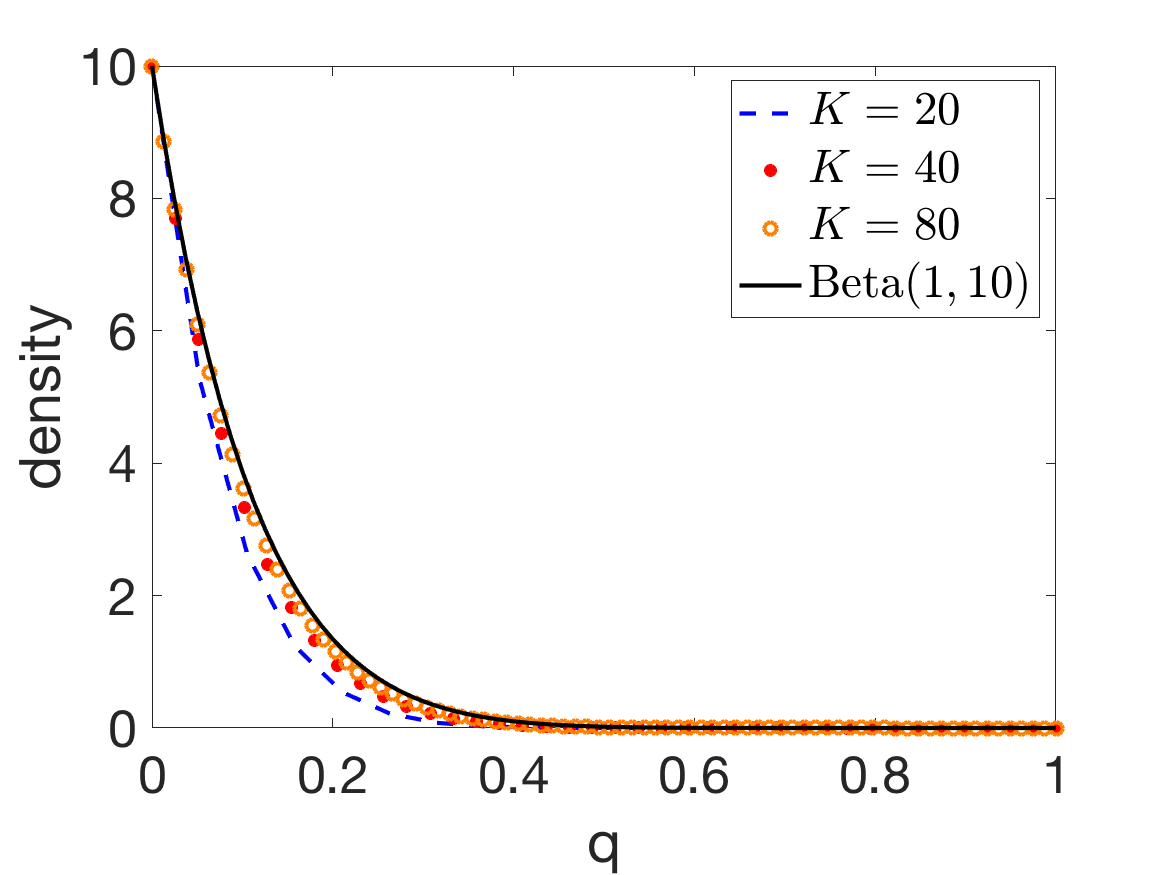}}
    \caption{\small{Figure shows the profiles of $K\gtKj$ (i.e., the probability distribution of the quantile $q=j/K$ for the RtG procedure, as described in the text) as a function of $q$. We consider three profiles (of  $K\gtKj$) for three different values of $K$, and the $\text{Beta}(1,10)$ density function (we fix $t=10$). We observe that for $K \approx 40$, the profile of $K\gtKj$ and that of the $\text{Beta}(1,10)$ distribution are almost indistinguishable.}
    }
    \label{fig:limit}
\end{figure}

Figure \ref{fig:limit} compares the probability distribution of the discrete random variable $j/K$ and its continuous limit: as soon as $K \approx 40$, the function $K\gtKj$ becomes (almost) identical to the Beta density.


Given a partition $\{I_g\}_{g\in [G]}$ of $[K]$, let us denote for every $g \in [G]$:
\begin{equation}\label{ugly-defnition}
b_g=\max_{j\in I_g} |a_j|~~~~ \text{and}~~~ k_g = \argmax_{j \in I_g} |a_j|\ .
\end{equation}
For the RtG procedure with Type 2 random selection rule, note that 
$\PP(\hj = k_g) =1/G$ for all $g \in [G]$. 
Type 3 selection rule is a combination of Type 1 and Type 2 selection rules. One can view the RtG procedure with Type 3 selection rule as a two-step procedure: (i) compute $b_{g}$ and $k_g$ as in~\eqref{ugly-defnition};
and (ii) use a RtG procedure with Type 1 rule on $\{b_g\}_{g\in[G]}$. Using an argument similar to that used in Proposition \ref{prop:prob-R1}, we have
\begin{equation}\label{eq:prob-T3}
\PP({\hj}= k_{(g)} ) = \gamma_t^G(g) \ ,   
\end{equation}
where we recall that $|a_{k_{(1)}}| \geq |a_{k_{(2)}}| \geq \ldots\geq |a_{k_{(G)}}|$ and 
$b_{(g)} =|a_{k_{(g)}}|$ for all $g$.

The following Proposition establishes a connection among the four types of selection rules and the four norms described in Section \ref{sec:three-norms}.

\smallskip

\begin{prop}\label{prop:norm-a}
Consider the RtG procedure for finding the approximate maximum of the absolute values of $a$. It holds that
$$
\EE [|a_{\hj}| ] = \|a\|_{\FF} \ ,
$$
where $\FF$ denotes infinity norm, the ordered $\ell_1$ norm with parameter $\gamma=[\gtKj]_j$, the $\ell_{1,\infty}$ group norm, or the ordered mixed norm with parameter $\gamma=[\gtGj]_j$ when the selection rule is Type 0, Type 1, Type 2 or Type 3 (cf Section \ref{sec:RGBM}), respectively.
\end{prop}
{\bf Proof.} 

{\bf Type 0:} This corresponds to the deterministic case and $|a_{\hj}|=\max_j|a_j|=\|a\|_{\infty}$.

{\bf Type 1:} It follows from Proposition \ref{prop:prob-R1} that $\PP(\hj=(j))=\gtKj$, thus
$$
\EE [|a_{\hj}|]  = \sum_{j=1}^K \gtKj |a_{(j)}| = \|a\|_{\mS}\   .
$$

{\bf Type 2:} For the Type 2 random selection rule,
we have $\PP({\hj}=k_g)=\tfrac{1}{G}$ for any $g\in [G]$, thus:
$$
\EE [|a_{\hj}|] = \frac{1}{G} \sum_{g=1}^G b_g =\frac{1}{G} \sum_{g=1}^G\|a_{I_g}\|_{\infty} = \|a\|_{\GG}\ .
$$

{\bf Type 3:} It follows from \eqref{eq:prob-T3} that
$$
\EE [|a_{\hj}|] =\sum_{g=1}^G \gtGg b_{(g)} = \sum_{g=1}^G \gtGg \|a_{I_{(g)}}\|_{\infty} = \|a\|_{\CC} \ . \qed
$$ 


\section{Computational Guarantees for RGBM}\label{sec:guarantees}
Here we derive computational guarantees for RGBM. We first introduce some standard regularity/continuity conditions on the scalar loss function $\ell(y,f)$ that we require in our analysis.

\smallskip

\begin{mydef}
We denote ${\partial\ell(y,f)}/{\partial f}$ as the derivative of the scalar loss function $\ell$ wrt the prediction $f$. We say that $\ell$ is $\sigma$-smooth if for any $y$ and predictions $f_1$ and $f_2$, it holds that
$$
\ell(y, f_1) \le \ell(y, f_2) + \frac{\partial \ell (y, f_2)}{\partial f}(f_1-f_2) + \frac{\sigma}{2} (f_1-f_2)^2 .
$$

We say $\ell$ is $\mu$-strongly convex (with $\mu>0$) if for any $y$ and predictions $f_1$ and $f_2$, it holds that
$$
\ell(y, f_1) \ge \ell(y, f_2) +  \frac{\partial \ell (y, f_2)}{\partial f}(f_1-f_2) + \frac{\mu}{2} (f_1-f_2)^2 .
$$
\end{mydef}

We present examples of some loss functions commonly used in GBM along with their 
regularity/continuity parameters:

{\bf squared $\ell_2$ or least squares loss:} $\ell(y,f)=\tfrac{1}{2}(y-f)^2$ is $1$-smooth and $1$-strongly convex.

{\bf Huber loss:} The Huber loss function with parameter $d>0$, given by
\begin{equation*}
    l_d(y,f)= \left\{\begin{array}{ll}
    \tfrac{1}{2}(y-f)^2     &\text{ for } |f-y|\le d  \\
    d|y-f|-\tfrac{1}{2} d^2     & \text{ otherwise },
    \end{array}\right.
\end{equation*}
is $1$-smooth but not strongly convex.

{\bf Logistic loss:} We consider a regularized version of the usual logistic loss function: $\ell_d(y,f)=\log(1+e^{-yf})+\frac{d}{2}f^2$
with $d \geq 0$, which is $(\frac{1}{4}+d)$-smooth and $d$-strongly convex (when $d>0$). A special case is the usual logistic loss when $d=0$, which is $\frac{1}{4}$-smooth but not strongly convex.

{\bf Exponential loss:} $\ell(y,f)=\exp(-yf)$ is neither strongly convex nor smooth.

\medskip

Notice that the objective function $\LC(\beta)$ has an invariant subspace in the coefficient space, namely for any $\omega\in \Ker(B)$, it holds that $\LC(\beta)=\LC(\beta+\omega)$. Let us denote
\begin{equation}\label{eq:subspace}
Z(\hbeta):=\left\{ \beta~|~B\beta=B\hbeta\right\}    
\end{equation}
as the invariant subspace of $\hbeta$. Recall that $\FF\in \{\infty, \mS, \GG, \CC\}$ and $\FF^*$ is the dual norm of $\FF$ (see Section~\ref{sec:three-norms}). We define a distance metric in the $\beta$-space as:
$$
\DistFB(\beta_1,\beta_2):=\DistF(Z(\beta_1),Z(\beta_2))=\min_{b\in Z(\beta_1),\hb\in Z(\beta_2)} \|b-\hb\|_{\FF^*}=\min_{\omega\in\Ker(B)} \|\beta_1-\beta_2-\omega\|_{\FF^*} \ ,
$$
which is the usual notion of distance between subspaces in the $\FF^*$ norm. In particular, if $\beta_1,\beta_2\in Z(\hbeta)$, then $\DistFB(\beta_1,\beta_2)=0$.
Note that $\DistFB$ is a pseudo-norm---Proposition~\ref{prop:basic-prop} lists a few properties of $\DistFB$.

\smallskip

\begin{prop} \label{prop:basic-prop} \ 
\begin{enumerate}
    \item $\DistFB(\beta_1, \beta_2)$ is symmetric: i.e., for any $\beta_1$ and $\beta_2$, we have
    $$
    \DistFB(\beta_1, \beta_2) = \DistFB(\beta_2, \beta_1) \ .
    $$

    \item $\DistFB(\beta_1, \beta_2)$ is translation invariant: i.e., for any $\beta_1$, $\beta_2$ and $\hbeta$, we have
    $$
    \DistFB(\beta_1-\hbeta,\beta_2-\hbeta)=\DistFB(\beta_1,\beta_2) \ .
    $$
\end{enumerate}
\end{prop}
{\bf Proof.} \ 

1. The proof of this part follows from
$$\DistFB(\beta_1, \beta_2)=\min_{b\in Z(\beta_1),\hb\in Z(\beta_2)} \|b-\hb\|_{\FF^*} = \min_{b\in Z(\beta_1),\hb\in Z(\beta_2)} \|\hb-b\|_{\FF^*}=\DistFB(\beta_2, \beta_1) \ .$$

2. The proof of this part follows from $$\DistFB(\beta_1-\hbeta,\beta_2-\hbeta)=\min_{\omega\in\Ker(B)} \|(\beta_1-\hbeta)-(\beta_2-\hbeta)-\omega\|_{\FF^*}=\min_{\omega\in\Ker(B)} \|\beta_1-\beta_2-\omega\|_{\FF^*}=\DistFB(\beta_1,\beta_2) \ .$$

\subsection{Minimal Cosine Angle}\label{sec:MCA}
Here we introduce a novel geometric quantity Minimal Cosine Angle (MCA) $\Theta_{\FF}$, which measures the density of the collection of weak-learners in the prediction space with respect to the $\FF$ norm. We show here that MCA plays a key role in the computational guarantees for RGBM.

\smallskip

\begin{mydef}
The Minimal Cosine Angle (MCA) of a set of weak-learners (given by the columns of the matrix $B$) with respect to the $\FF$ norm is defined as:
\begin{equation}\label{eq:Theta}
    \Theta_{\FF}:=\min_{c\in\emph{\Range(B)}}\left\|\left[\cos(\Bj,c)\right]_{j=1,\ldots,K}\right\|_{\FF}.
\end{equation}
\end{mydef}


\textcolor{black}{\begin{rem}
At first look, the MCA quantity seems to be similar to the Cheung-Cucker condition number for solving a linear system. 
See~\cite{cheung2001new,epelman2000condition} for details on the Cheung-Cucker condition number, and \cite{ramdas2014margins,ramdas2016towards} for how it connects to margins and the convergence rate for iterative algorithms (e.g., the Perceptron and Von-Neumann algorithms) arising in binary classification tasks.  However, there is an important basic difference: Our measure MCA looks at the {\it columns} of the basis matrix $B$ whereas the Cheung-Cucker condition number is based on the {\it rows} of $B$. 
\end{rem}}

\begin{figure}[!tbp]
\centering
  \begin{subfigure}[b]{0.33\textwidth}
    \includegraphics[width=\textwidth, trim=0 0 0 0, clip]{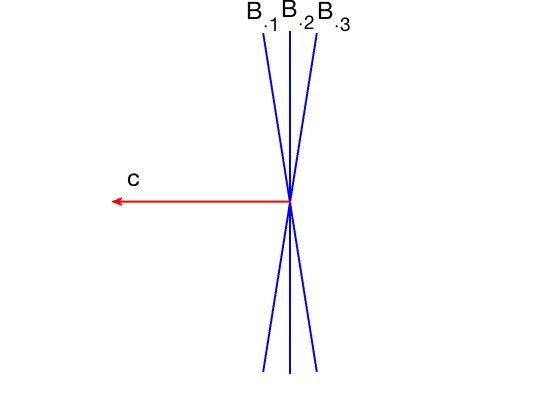}
    \caption{Poorly spread}
  \end{subfigure}
  \hspace*{-0.7em}
  \begin{subfigure}[b]{0.33\textwidth}
    \includegraphics[width=\textwidth, trim=0 0 0 0, clip]{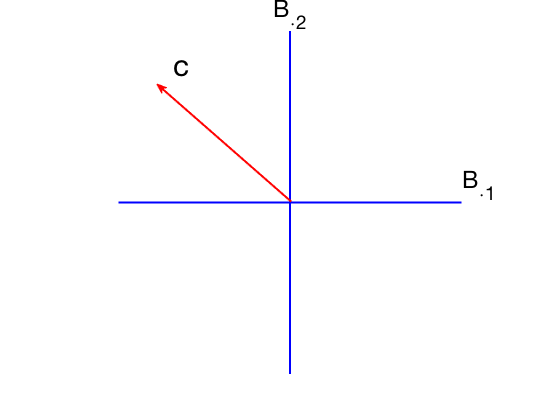}
    \caption{Moderately spread}
  \end{subfigure}
  \hspace*{-0.7em}
  \begin{subfigure}[b]{0.33\textwidth}
    \includegraphics[width=\textwidth, trim=0 0 0 0, clip]{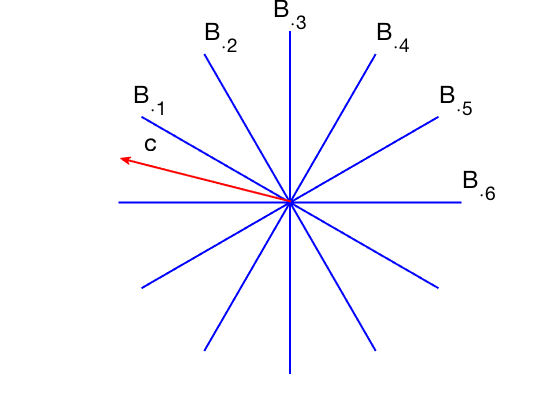}
    \caption{Densely spread}
  \end{subfigure}
  \caption{\textcolor{black}{Illustration of the relationship between $\Theta_{\infty}$ and density of weak-learners in a $2$D example. Figures in panels (a), (b) and (c) represent poorly-spread weak-learners, moderately spread weak-learners and densely spread} weak-learners, respectively. When $\FF$ is the infinity norm, the values of $\Theta_{\infty}$ are given by: (a) $\Theta_{\infty}^2\approx 0$; (b) $\Theta_{\infty}^2= 1/2$; and (c) $\Theta_{\infty}^2\approx 0.933$---the weak-learners are more spread out for higher values of $\Theta_{\infty}$.}
  \label{fig:spread}
\end{figure}

The quantity $\Theta_{\FF}$ measures the ``density'' of the weak-learners in the prediction space. Figure \ref{fig:spread} provides an illustration in a simple $2$D example when $\FF$ is the infinity norm. Given weak-learners 
$B_{: 1},\ldots,B_{:K}$, we compute the cosine of the angle between each weak-learner and a direction $c$. The $\FF$ norm can be viewed as an approximation to the infinity norm, which is the norm corresponding to traditional GBM. The quantity MCA refers to the minimum (over all directions indexed by $c$) of such reweighted angles.

We next present some equivalent definitions of $\Theta_{\FF}$:

\smallskip

\begin{prop}\label{prop:norm-transformation}

\begin{equation}\label{eq:norm-trans}
\Theta_{\FF}=\min_{c\in\emph{\Range}(B)}\frac{\|B^{T}c\|_{\FF}}{\|c\|_{2}}=\min_{a}\frac{\|Ba\|_{2}}{\DistFB(0,a)} > 0\ .    
\end{equation}

\end{prop}

\textbf{Proof.} 
{\color{black}
The first equality follows directly by rewriting \eqref{eq:Theta}. Notice that for any norm $\FF$ in $\RR^K$ (a finite dimensional space), there exists a scalar $\gamma>0$ such that $\|B^Tc\|_{\FF}\ge\gamma\|B^Tc\|_{\infty}$. Thus
$$
\Theta_{\FF} =\min_{c\in\Range(B), \|c\|_2=1}\|B^{T}c\|_{\FF}\ge \gamma \|B^{T}c\|_{\infty}>0\ ,
$$
where the second inequality follows from the observation that $c\in\Range(B)$. We now proceed to show the second equality of \eqref{eq:norm-trans}. 

By the definition of $\DistFB$ and the definition of the dual norm, we have
\begin{equation*}\label{eq:strong-0}
\begin{array}{lcl}
    \DistFB(0,a) & =&\displaystyle\min_{\omega\in\Ker(B)}\|a-\omega\|_{\FF^*}=\min_{\omega\in\Ker(B)}\max_{\|b\|_{\FF}\le 1}\langle a-\omega, b\rangle = \max_{\|b\|_{\FF}\le 1}\min_{\omega\in\Ker(B)}\langle a-\omega, b\rangle \\ \\
     & = &\displaystyle \max_{\|b\|_{\FF}\le 1, b\in\Range(B^T)}\langle a, b\rangle =\max_{\|b\|_{\FF}\le 1, b\in\Range(B^T)} |\langle a, b\rangle| = \max_{b\in\Range(B^T)}\frac{|\langle a, b\rangle|}{\|b\|_{\FF}} \ ,
\end{array}
\end{equation*}
where the third equality uses Von Neumann's Minimax Theorem, and the fourth equality is based on the observation
$$
\min_{\omega\in\Ker(B)}\left\langle a-\omega,b\right\rangle =\left\{ \begin{array}{cl}
-\infty &\text{ for }b\not\in\Range(B^{T})\\
\langle a,b\rangle &\text{ for }b\in\Range(B^{T}) \ .
\end{array}\right.
$$
Therefore,
\begin{equation*}
\min_{a}~\frac{\|Ba\|_{2}}{\DistFB(0,a)}=\min_{b\in\Range(B^{T}),a}~\frac{\|Ba\|_{2}\|b\|_{\FF}}{\left|\langle a,b\right\rangle| }\ .
\end{equation*}
Denote $P_{B}=B^{T}(BB^{T})^{\dagger}B$ as the projection matrix onto $\Range(B^{T})$, then we have $P_{B}b=b$ for any $b\in\Range(B^{T})$. Thus
\begin{equation}
    \min_{a}~\frac{\|Ba\|_{2}}{\DistFB(0,a)}= \min_{b\in\Range(B^{T}),a}~\frac{\|Ba\|_{2}\|b\|_{\FF}}{|\left\langle a,P_{B}b\right\rangle| } = \min_{b\in\Range(B^{T}),a}~\frac{\|Ba\|_{2}\|b\|_{\FF}}{
|\left\langle Ba,(BB^{T})^{\dagger}Bb\right\rangle|} \ .
\label{eq:strong_1}
\end{equation}

Now denote $c=(BB^{T})^{\dagger}Bb$, then $c\in\Range(B)$ and $B^{T}c=P_{B}b=b$.
Note that for any $a$, we have $
\|Ba\|_{2}\|c\|_{2}\ge\left|\left\langle Ba,c\right\rangle \right|,
$
which implies
\[
\min_{a}~\frac{\|Ba\|_{2}}{|\left\langle Ba,c\right\rangle| }\ge\frac{1}{\|c\|_{2}}\ .
\]
Since $c\in\Range(B)$, there exists a vector $a$ satisfying $Ba=c$, which leads to 
\[
\frac{\|Ba\|_{2}}{|\left\langle Ba,c\right\rangle| }=\frac{\|c\|_{2}}{\|c\|_{2}^{2}}=\frac{1}{\|c\|_{2}}\ ,
\]
from which it follows that
\begin{equation}
\min_{a}~\frac{\|Ba\|_{2}}{|\left\langle Ba,c\right\rangle| }=\frac{1}{\|c\|_{2}}\ .\label{eq:middle-proposition-1}
\end{equation}
Substituting $c=(BB^{T})^{\dagger}Bb$ and combining \eqref{eq:strong_1} and \eqref{eq:middle-proposition-1}
 yields
\[
\min_{a}~\frac{\|Ba\|_{2}}{\DistFB(0,a)}=\min_{c\in\Range(B)}~\frac{\|B^{T}c\|_{\FF}}{\|c\|_{2}}\ ,
\]
which completes the proof. \qed}

To gain additional intuition about MCA, we consider some examples:

{{\bf Example 1} (Orthogonal Basis with Infinity Norm)} Suppose $\FF$ is the infinity norm and the set of weak-learners in $\RR^p$ forms an orthogonal basis (e.g, the discrete Fourier basis in $\RR^p$), then $\Theta_{\infty}={1}/{\sqrt{p}}$.

{{\bf Example 2} (Orthogonal Basis with Ordered $\ell_1$ Norm)} Suppose $\FF$ is the ordered $\ell_1$ norm with a parameter sequence $\gamma\in\RR^p$ and the set of weak-learners in $\RR^p$ forms an orthogonal basis, then
\begin{equation}\label{eq:theta-slope}
\Theta_{\mS}=\min\left\{\gamma_1, \frac{1}{\sqrt{2}}(\gamma_1+\gamma_2),\ldots,\frac{1}{\sqrt{p}}(\gamma_1+\ldots+\gamma_p)\ \right\} \ .    
\end{equation}

We present a proof for~\eqref{eq:theta-slope}---note that the result for Example~1 follows as a special case. Without loss of generality, we assume $B$ to be an identity matrix. {It then follows from the second equality of \eqref{eq:norm-trans} that}
\begin{equation}\label{eq:OB-1}
    \Theta_{\mS} = \min_{\|a\|_{\mS^*}= 1} \|a\|_2 \ .
\end{equation}
By flipping the constraint and the objective function of \eqref{eq:OB-1} we can instead consider the following equivalent problem:
\begin{equation*}
    \Phi = \max_{\|a\|_2= 1} \|a\|_{\mS^*}=\max_{\|a\|_2\le 1} \|a\|_{\mS^*}\ ,
\end{equation*}
and we have $\Theta_{\mS}={1}/{\Phi}$. Using the definition of the dual of the ordered $\ell_1$ norm (see \eqref{eq:dual-slope}), notice that for any $i\in [p]$, it follows from the $\ell_1$-$\ell_2$ norm inequality that 

$$\sum_{j=1}^i |a_{(j)}| \le \sqrt{i \left(\sum_{j=1}^i a_{(j)}^2\right)}\le \sqrt{i}\|a\|_2\le \sqrt{i}\ , $$

and therefore
\begin{equation*}
    \Phi = \max_{\|a\|_2\le 1} \max_{i\in [p]}\left\{ \frac{\sum_{j=1}^i |a_{(j)}|}{\sum_{j=1}^i \gamma_j}\right\} \le  \max_{i\in [p]}\left\{ \frac{\sqrt{i}}{\sum_{j=1}^i \gamma_j}\right\} \ .
\end{equation*}
For any $i\in[p]$ define $\ta_1=\cdots=\ta_i={1}/{\sqrt{i}}$ and $\ta_{i+1}=\cdots=\ta_p=0$, then we have $\Phi\ge\|\ta\|_{\mS^*}= {\sqrt{i}}/({\sum_{j=1}^i \gamma_j})$. Therefore, we have:
$$\Phi=\max_{i\in [p]}\left\{ \frac{\sqrt{i}}{\sum_{j=1}^i \gamma_j}\right\}=\frac{1}{\Theta_{\mS}}$$
which completes the proof of~\eqref{eq:theta-slope}.


\smallskip
\smallskip

\begin{figure}
    \centering
    \includegraphics[width=0.4\textwidth, trim=0 0.5 0 0.5, clip=true]{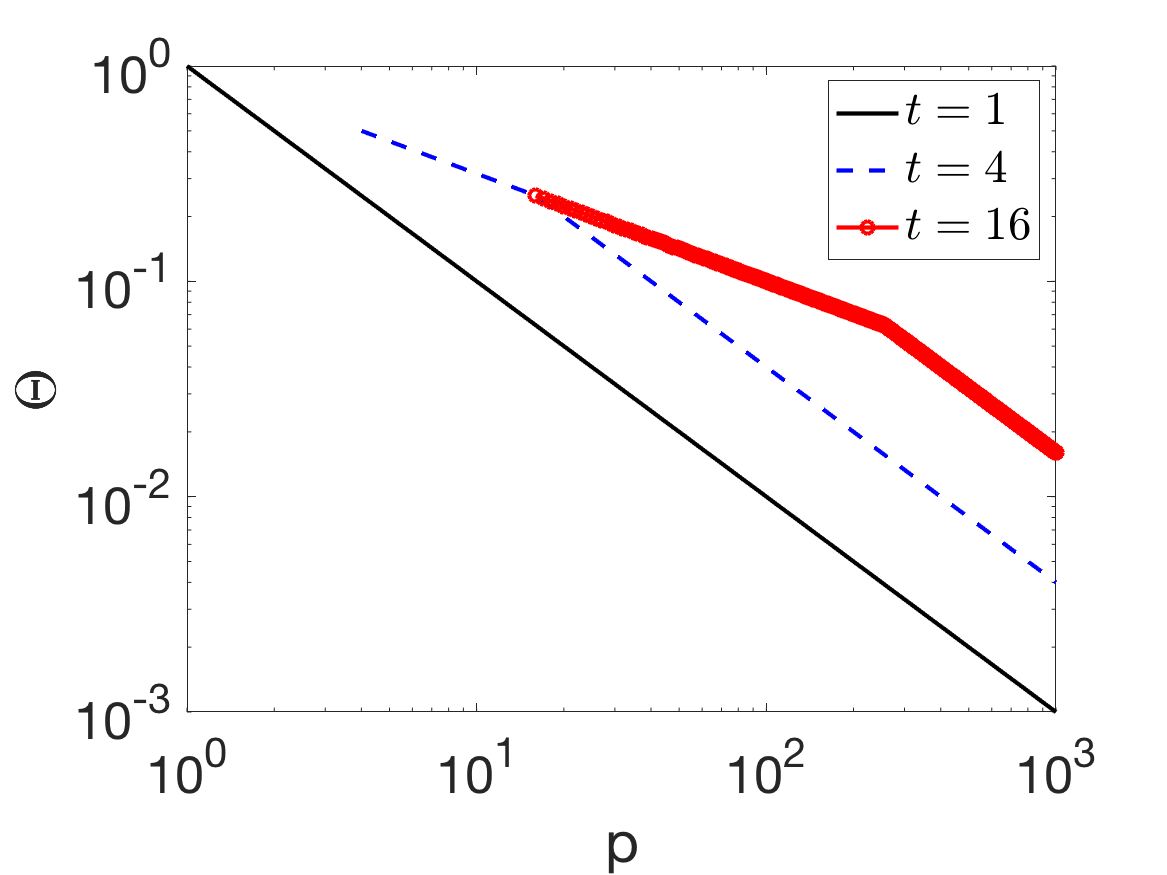}
    \caption{\small{Plot shows how $\Theta_{\mS}$ varies with $p$ (log-log plot) when the weak-learners are orthogonal and $\FF$ corresponds to the ordered $\ell_1$ norm with parameter $\gamma=[\gamma_{t}^p(j)]_j$ (see~\eqref{eq:gtKj}). We show three profiles for three different values of $t$.
    Note that $\Theta_{\mS}$ is defined only for $p \geq t$. The setup is described in Remark~\ref{rem:slope}.
    } }
    \label{fig:Theta}
\end{figure}

\begin{rem}\label{rem:slope}
Consider using a Type 1 random selection rule in RGBM, then the corresponding norm $\FF$ is the ordered $\ell_1$ norm with parameter $\gamma_t^p=[\gamma_t^p(j)]_j$ as defined in \eqref{eq:gtKj}. Figure \ref{fig:Theta} shows the value of $\Theta_{\mS}$ (computed by formula \eqref{eq:theta-slope}) versus the dimension $p$---we consider different values of $t$ and use an orthogonal basis. 
The figure suggests that $\Theta_{\mS}$ depends upon $p,t$ as follows:
$$
\Theta_{\mS} \sim \left\{ \begin{array}{cc}
     \frac{1}{\sqrt{p}}& \text{ if } p\le t^2,\\
     \frac{t}{p}& \text{ otherwise}.
\end{array}\right.
$$
\end{rem}

{{\bf Example 3} (Binary Basis with Infinity Norm)} Suppose $\FF$ is the infinity norm, and the basis matrix $B$ has entries $B_{i,j}\in \{-1,0,1\}$---leading to 
$3^p$ different weak-learners. In this case,
\begin{equation}\label{mca-value-eg-3}
\Theta_{\infty} = \frac{1}{\sqrt{1^2+(\sqrt{2}-1)^2+\cdots+(\sqrt{p}-\sqrt{p-1})^2}}\propto \frac{1}{\sqrt{\ln p}} \ .
\end{equation}
We present a proof for~\eqref{mca-value-eg-3}. 
Since $B_{i,j}\in\{-1,0,1\}$, we have:
\begin{equation}
    \Theta_{\infty} = \min_c \max_j |\cos(\Bj,c)|=\min_c \max_{i\in[p]} \max_{\|\Bj\|_1=i} |\cos(\Bj,c)|=\min_{c} \max_{i\in[p]} \frac{\sum_{k=1}^i |c_{(k)}|}{\sqrt{i}\|c\|_2} \ .
\end{equation}
Let us recall the form of $\mS^*$ i.e., the dual ordered $\ell_1$ norm 
appearing in Proposition~\ref{prop:dual-norm}. Observe that $\max_{i\in[p]} (\sum_{k=1}^i |c_{(k)}|/\sqrt{i})= \|c\|_{\mS^*}$, where $\gamma=[\sqrt{i}-\sqrt{i-1}]_{i\in [p]}$ is the parameter of the ordered $\ell_1$ norm $\mS$. Thus $$\Theta_{\infty}=\min_{c}~\frac{\|c\|_{\mS^*}}{\|c\|_2}=\min_{a}~\frac{\|a\|_2}{\|a\|_{\mS}}=\min_{\|a\|_{\mS}=1}~\|a\|_2 \ $$ where, the second equality uses  \eqref{eq:norm-trans} with $\FF=\mS^*$ and $B$ as the identity matrix. By flipping the constraint and the objective function, we can instead consider the following equivalent problem:
$$
\Phi = \max_{\|a\|_2=1} \|a\|_{\mS}=\max_{\|a\|_2\le 1} \|a\|_{\mS} \ ,
$$
with $\Theta_{\infty}= {1}/{\Phi}$. By the Cauchy-Schwarz inequality, it holds that
\begin{equation*}
    \|a\|_S^2 = \left(\sum_{i=1}^p \gamma_i |a_{(i)}|\right)^2\le \left(\sum_{i=1}^p \gamma_i^2\right)\|a\|_2^2 =\left(\sum_{i=1}^p\left(\sqrt{i}-\sqrt{i-1}\right)^2\right)\|a\|_2^2\ ,
\end{equation*}
\textcolor{black}{with equality being achieved when $a\propto[\sqrt{i}-\sqrt{i-1}]_i$.} Thus we have $\Phi=\sqrt{\sum_{i=1}^p\left(\sqrt{i}-\sqrt{i-1}\right)^2}$ and
$\Theta_{\infty}=1/\Phi$.
Notice that
\begin{equation*}
    \frac{1}{4}\sum_{i=1}^p\frac{1}{i}\le \sum_{i=1}^p \left(\sqrt{i}-\sqrt{i-1}\right)^2=\sum_{i=1}^p\frac{1}{\left(\sqrt{i}+\sqrt{i-1}\right)^2}\le 1 +\frac{1}{4}\sum_{i=2}^p\frac{1}{i-1} \ ,
\end{equation*}
where the lhs and rhs of the above are both $O(\ln p)$.
This implies that
$\sum_{i=1}^p \left(\sqrt{i}-\sqrt{i-1}\right)^2 \propto \ln p,$
thereby completing the proof.
\qed


\smallskip
\smallskip

\begin{rem}
The binary basis described in Example~3 (with $\Theta_{\infty} = O({1}/{\sqrt{\ln p}})$) is more densely distributed in the prediction space when compared to Example~1 (with $\Theta_{\infty}=O({1}/{\sqrt{p}})$)---See Figure~\ref{fig:spread} (b) and (c) for a schematic illustration. 
\end{rem}





\subsection{Computational Guarantees: Strongly Convex Loss Function}
We establish computational guarantees for RGBM when the scalar loss function $\ell$ is both smooth and strongly convex. Let $\EE_{m}$ denote the expectation over the random selection scheme at iteration $m$, conditional on the selections up to iteration $m-1$. Let $\EE_{\xi_m}$ denote the expectation over the random selection scheme up to (and including) iteration $m$. The following theorem presents the linear convergence rate for RGBM.

\smallskip

\begin{thm}\label{thm:strong}
Let $\ell$  be $\mu$-strongly convex and $\sigma$-smooth. Consider RGBM (Algorithm \ref{al:rgbm}) or RtGCD (Algorithm \ref{al:rgcd}) with either a line-search step-size rule or constant step-size rule with $\rho={1}/{\sigma}$. If $\Theta_{\FF}$ denotes the value of the corresponding MCA, then for all $M \geq 0$ we have:
\begin{equation}\label{eq:strong_result}
\E_{\xi_{M}}[\LC(\beta^{M})-\LC(\beta^{*})]\le \left(1-\tfrac{\mu}{\sigma}\Theta_{\FF}^2\right)^{M}\left(\LC(\beta^{0})-\LC(\beta^{*})\right). 
\end{equation}
\end{thm}
Notice that in the special case when $J$ is chosen deterministically as the set of all weak-learners, Theorem \ref{thm:strong} leads to a linear convergence rate for GBM~\cite{friedman2001greedy}. Some prior works have also presented a linear convergence rate for GBM, but our results are different. For example, \cite{telgarsky2012primal} shows a linear convergence rate but the constant is exponential in the number of features $p$, except for the exponential loss\footnote{The result of~\cite{telgarsky2012primal} for the exponential loss function is superior to that presented here, as their analysis is targeted towards this loss function.}. \cite{freund2017new} presents a linear convergence rate for LS-Boost (GBM with a least squares loss function) of the form $O(\tau^M)$, where  
the parameter $\tau=1-\lambda_{\text{pmin}}(B^TB)/4K$ depends upon $\lambda_{\text{pmin}}(A)$, the minimal non-zero eigenvalue of a matrix $A$. In GBM, $K$ is usually exponentially large, thus $\tau$ can be close to one. 
The linear convergence constant derived herein (i.e. $1-\tfrac{\mu}{\sigma}\Theta_{\FF}^2$), has a superior dependence on the number of weak-learners, and it stays away from $1$ as $K$ becomes large.
We obtain an improved rate since we employ a different analysis technique based on MCA.

\smallskip

\begin{rem}
We study the convergence rate of RGBM as a function of $t$ using the same setup considered in Remark \ref{rem:slope}. 
Using an ``epoch'' (i.e., the cost to evaluate all weak-learners across all samples) as the unit of computational cost, the cost per iteration of RGBM is $t/p$ epochs. Then the (multiplicative) improvement per epoch is
\begin{equation*}
    \left(1-\frac{\mu}{\sigma}\Theta_{\mS}^2\right)^{p/t}\sim\left\{\begin{array}{cc}
    \left(1-\frac{\mu}{p\sigma}\right)^{p/t}     & \text{if  } t\ge\sqrt{p},\\
    \left(1-\frac{t^2\mu}{p^2\sigma}\right)^{p/t}     & \text{otherwise}.
    \end{array}\right.
\end{equation*}
This suggests that we should choose $t\sim \sqrt{p}$ when the weak-learners are almost orthogonal. Recall that from a coordinate descent perspective, RtGCD with $t=1$ leads to random CD, and RtGCD with $t=p$ leads to greedy CD. 
Choosing $t$ to be larger than $O(\sqrt{p})$ will not lead to any improvement in the theoretical convergence rate, though it will lead to an increase in computational cost.
\end{rem}


{\color{black}
\begin{rem}
Since traditional GBM is equivalent to greedy CD in the coefficient space,  theoretical guarantees of greedy CD can be used to analyze GBM. In this case however, the resulting computational guarantees may contain the total number of weak learners $K$ --- the bounds we present here are of a different flavor (they depend upon MCA). \\
Recently, interesting techniques have been proposed to improve the efficiency of greedy CD. For example, \cite{stich2017approximate} propose a scheme to approximate the entire gradient vector and use it to update the coordinates (in the spirit of approximate steepest CD).
It will be interesting to adapt the ideas from~\cite{stich2017approximate} to the case of Boosting-like algorithms presented herein.
\end{rem}}

Propositions~\ref{prop:expectation}--\ref{prop:strong_con} presented below will be needed for the proof of Theorem \ref{thm:strong}. 
Proposition~\ref{prop:expectation} establishes a relationship among the four selection rules for choosing subset $J$ in RGBM (Algorithm \ref{al:rgcd}) and the norms introduced in 
Section~\ref{sec:three-norms}.

\smallskip

\begin{prop}\label{prop:expectation}
Consider Algorithm \ref{al:rgcd} with the four types of selection rules for choosing the set $J$ as described in Section \ref{sec:rules-select}. For any iteration index $m$, we have
$$
\EE_{m} \left[(\nabla_{j_m} \LC(\beta^m))^2\right] = \left\|\left[\nabla_j \LC(\beta^m)^2\right]_j\right\|_{\FF}\ge \|\nabla \LC(\beta^m)\|_{\FF}^2\ ,
$$
where $\FF$ is the infinity norm, the ordered $\ell_1$ norm with parameter $\gamma=[\gtKj]_j$, the $\ell_{1,\infty}$ group norm, or the ordered mixed norm with parameter $\gamma=[\gtGj]_j$ when the selection rule is Type 0, Type 1, Type 2 or Type 3, respectively.
\end{prop}


{\bf Proof.}  The equality is a direct result of Proposition \ref{prop:norm-a} with $a_j=(\nabla_{j} \LC(\beta^m))^2$. Notice that the $\FF$ norm of $a$ is a weighted sum of its coordinates---for notational convenience, we denote these weights by a vector $\lambda\in \mathbb{R}^K$ that satisfies: $\left\|\left[\nabla \LC_j(\beta^m)^2\right]_j\right\|_{\FF} = \sum_j \lambda_j\left( \nabla_j\LC(\beta^m)\right)^2$; and $\lambda_{j}\ge 0, j \in [K]$, $\sum_{j} \lambda_j=1$.
Thus we have  
$$
\left\|\left[\nabla \LC_j(\beta^m)^2\right]_j\right\|_{\FF}=\left(\sum_{j} \lambda_j\right)\left(\sum_j \lambda_j\left( \nabla_j\LC(\beta^m)\right)^2\right) \ge\left(\sum_j  \lambda_j\left| \nabla_j\LC(\beta^m)\right|\right)^2 = \|\nabla \LC(\beta^m)\|_{\FF}^2\ ,
$$
where the inequality above, follows from the Cauchy Schwarz inequality.
\qed

\smallskip

The following proposition can be viewed as a generalization of the mean-value inequality.

\smallskip

\begin{prop}\label{prop:average}
For $a\in\Range(B^{T})$ and $t>0$, it holds that
$$
\min_{\beta}\left\{\left\langle a,\beta-\beta^{*}\right\rangle +\frac{t}{2}\DistFB(\beta,\beta^{*})^2\right\}=-\frac{1}{2t}\|a\|_{\FF}^{2}\ .
$$
\end{prop}
{\bf Proof:} Let $b=\beta-\beta^{*}$, $\Ker(B)=\left\{ \omega~|~B\omega=0\right\} $
and $c=b+\omega$. By the definition of $\DistFB$, we have:
\begin{equation*}
\begin{array}{cl}
 &\displaystyle \min_{\beta}\left\{\left\langle a,\beta-\beta^{*}\right\rangle +\frac{t}{2}\DistFB(\beta,\beta^{*})^2\right\}= \displaystyle \min_{b}\min_{\omega\in\Ker(B)}\left\{\left\langle a,b\right\rangle +\frac{t}{2}\|b+\omega\|_{\FF^{*}}^{2}\right\}
\\ \\
= &\displaystyle \min_{\omega\in\Ker(B)}\left\{-\left\langle a,\omega\right\rangle +\min_{b+\omega}\left\langle a,b+\omega\right\rangle +\frac{t}{2}\|b+\omega\|_{\FF^{*}}^{2}\right\} = \displaystyle \min_{\omega\in\Ker(B)}\min_{c}\left\{\left\langle a,c\right\rangle +\frac{t}{2}\|c\|_{\FF^{*}}^{2}\right\}\\ \\
= &\displaystyle \min_{c}\left\{\left\langle a,c\right\rangle +\frac{t}{2}\|c\|_{\FF^{*}}^{2}\right\}\\
\end{array}
\end{equation*}
where the third equality considers $a\in\Range(B^{T})$ and makes use of the observation that $\left\langle a,\omega\right\rangle =0$
for $\omega\in\Ker(B)$.
Notice that
$$
\frac{t}{2}\|c\|_{\FF^{*}}^{2}+\frac{1}{2t}\|a\|_{\FF}^{2}\ge\|c\|_{\FF^{*}}\|a\|_{\FF}\ge|\left\langle a,c\right\rangle|
$$
and hence: $\min_{c}\left\{\left\langle a,c\right\rangle +\frac{t}{2}\|c\|_{\FF^{*}}^{2}\right\}\le-\tfrac{1}{2t}\|a\|_{\FF}^{2}$.
Now, if $\hc=\tfrac{1}{t}\|a\|_{\FF}\argmin_{\|c\|_{\FF^{*}}\le1}\left\langle a,c\right\rangle$,
we have 
$$\|\hc\|_{\FF^{*}}=\tfrac{1}{t}\|a\|_{\FF}~~~~\text{and}~~~~~\left\langle a,\hat{c}\right\rangle =-\tfrac{1}{t}\|a\|_{\FF}\max_{\|c\|_{\FF^{*}}\le1}\left\langle a,c\right\rangle =-\tfrac{1}{t}\|a\|_{\FF}^{2}\ ,$$
whereby $\left\langle a,\hc\right\rangle +\frac{t}{2}\|\hc\|_{\FF^{*}}^{2}=-\frac{1}{2t}\|a\|_{\FF}^{2}$.
Therefore it holds that
\[
\min_{c}\left\{\left\langle a,c\right\rangle +\tfrac{t}{2}\|c\|_{\FF^{*}}^{2}\right\}=-\tfrac{1}{2t}\|a\|_{\FF}^{2}\ ,
\]
which completes the proof. \qed

\smallskip
\smallskip

\begin{prop}\label{prop:strong_con}
If $\ell$ is $\mu$-strongly convex, it holds for any $\beta$ and $\hbeta$ that
$$
\LC(\hbeta)\ge\LC(\beta)+\left\langle \nabla\LC(\beta),\hbeta-\beta\right\rangle +\frac{1}{2}\mu\Theta_{\FF}^2\DistFB(\hbeta,\beta)\ .
$$
\end{prop}
{\bf Proof.} Since $\ell$ is $\mu$-strongly convex, we have
\begin{equation}
    \begin{array}{cl}
    \LC(\hbeta)     & = \displaystyle\sum_{i=1}^n \ell(y_i, B_{i:} \hbeta)  \\ \\
         & \ge \displaystyle\sum_{i=1}^n \left\{\ell(y_i, B_{i:} \beta) +  \frac{\partial \ell(y_i, \Bi \hbeta)}{\partial f} \langle B_{i:}, \hbeta_i - \beta_i\rangle + \tfrac{\mu}{2} \|B_{i:}\|_2^2(\hbeta_i - \beta_i)^2 \right\}\\ \\
         & = \LC(\beta) + \langle\nabla \LC(\beta), \hbeta-\beta\rangle + \tfrac{\mu}{2} \|B(\hbeta-\beta)\|_2^2 \\ \\
         & \ge \LC(\beta) + \langle\nabla \LC(\beta), \hbeta-\beta\rangle + \tfrac{\mu\Theta_{\FF}^2}{2} \DistFB(0, \hbeta-\beta)^2 \\ \\
         & = \LC(\beta) + \langle\nabla \LC(\beta), \hbeta-\beta\rangle + \tfrac{\mu\Theta_{\FF}^2}{2} \DistFB(\hbeta, \beta)^2 \ ,
    \end{array}
\end{equation}
where the second inequality follows from Proposition \ref{prop:norm-transformation}, and the last equality utilizes the symmetry and translation invariance of $\DistFB$ (Proposition \ref{prop:basic-prop}). \qed

\noindent \textbf{Proof of Theorem \ref{thm:strong}:} For either the line-search step-size rule or the constant step-size rule, it holds that
\begin{equation}\label{eq:thm_eq_1-1}
\begin{array}{cl}
\LC(\beta^{m+1}) & \le\LC(\beta^{m}-\tfrac{1}{\sigma}\nabla_{j_{m}}\LC(\beta^{m})e_{j_{m}})\\ \\
& \le L(\beta^m) -\frac{1}{\sigma}\nabla_{j_m} L(\beta^m) \langle\nabla L(\beta^m), e_{j_m}\rangle + \frac{1}{2\sigma} \|\nabla_{j_m}L(\beta^m) e_{j_m}\|^2 \\ \\
& = L(\beta^m) - \frac{1}{\sigma} \left(\nabla_{j_{m}}\LC(\beta^{m})\right)^{2} + \frac{1}{2\sigma} \left(\nabla_{j_{m}}\LC(\beta^{m})\right)^{2}\\ \\
& = L(\beta^m) - \frac{1}{2\sigma}\left(\nabla_{j_{m}}\LC(\beta^{m})\right)^{2} \ ,
\end{array}
\end{equation}
where the second inequality uses the fact that the loss function $\ell$ is $\sigma$-smooth. Thus $\LC(\beta^{m+1})\le\LC(\beta^{m})$ with probability one. 
As a result of Proposition \ref{prop:expectation}, taking expectation
over both sides of \eqref{eq:thm_eq_1-1} with respect to $\mathbb{E}_{m+1}$ yields
\begin{equation}\label{eq:decay}
   \mathbb{E}_{m+1}[\LC(\beta^{m+1})]\le\LC(\beta^{m})-\tfrac{1}{2\sigma}\|\nabla\LC(\beta^{m})\|_{\FF}^{2}\ . 
\end{equation}

Meanwhile, it follows from Proposition \ref{prop:strong_con} that

\begin{equation}\label{eq:strong_thm}
\begin{array}{lcl}
\LC(\beta^{*}) & =&\displaystyle\min_{\beta}\LC(\beta)\\ \\
 & \ge &\displaystyle\min_{\beta}\left[\LC(\beta^{m})+\left\langle \nabla\LC(\beta^{m}),\beta-\beta^{m}\right\rangle +\tfrac{\mu\Theta_{\FF}^{2}}{2}\DistFB(\beta,\beta^{m})\right]\\ \\
 & =&\LC(\beta^{m})-\tfrac{1}{2\mu\Theta_{\FF}^{2}}\|\nabla\LC(\beta^{m})\|_{\FF}^{2}\ ,
\end{array}
\end{equation}
where the last equality utilizes Proposition \ref{prop:average}. Note that~\eqref{eq:strong_thm} together with \eqref{eq:decay} leads to
\begin{equation*}
\mathbb{E}_{m+1}[\LC(\beta^{m+1})]-\LC(\beta^{*}) \le\LC(\beta^{m})-\LC(\beta^{*})-\frac{1}{2\sigma}\|\nabla\LC(\beta^{m})\|_{\FF}^{2} \le(1-\tfrac{\mu}{\sigma}\Theta_{\FF}^{2})(\LC(\beta^{m})-\LC(\beta^{*})) \ ,
\end{equation*}
and finally~\eqref{eq:strong_result} follows by a telescoping argument. \qed

\subsection{Computational Guarantees: Non-Strongly Convex Loss Function}

Define the initial level set of the loss function in 
the $\beta$-space (i.e., coefficient space) as
$$\LS_{0}=\left\{ \beta~|~\LC(\beta)\le\LC(\beta^{0})\right\},$$ and its maximal distance to the optimal solution set in $\DistFB$ as: $$\D=\max_{\beta\in\LS_{0}}\DistFB(\beta,\beta^{*})\ .$$
Note that $\LS_{0}$ is unbounded if $Z(\beta^0)$ (cf equation~\eqref{eq:subspace}) is unbounded. But interestingly, $\LS_{0}$ is bounded in $\DistF$, i.e. $\D<\infty$, when the scalar loss function $\ell$ has a bounded level set.

\smallskip

\begin{prop}\label{prop:bound}
Suppose $\ell$ has a bounded level set, then $\D$ is finite.
\end{prop}

\noindent {\bf Proof.} Since the convex function $\ell$ has a bounded level set, the set  $\{B(\beta-\beta^*)~|~\beta\in\LS_0\}$ is bounded. 
Thus there is a finite constant $C$ such that $\max_{\beta\in \LS_0}\|B(\beta-\beta^*)\|_2\le C$. Therefore,

\begin{equation*}
\begin{array}{lcl}
    \D & = &\displaystyle \max_{\beta\in\LS_0} \DistFB(0, \beta-\beta^*) \\ \\
     & \le &\displaystyle \max_{\|B(\beta-\beta^*)\|_2\le C} \DistFB(0, \beta-\beta^*) \\ \\
     & = &\displaystyle \max_{\|Ba\|_2\le C} \DistFB(0, a) \\ \\
     & \le &\displaystyle \max_{\|Ba\|_2\le C} \frac{\|Ba\|_2}{\Theta_{\FF}} \\ \\
     & = &\displaystyle \frac{C}{\Theta_{\FF}} \ ,
\end{array}
\end{equation*}
where the second inequality follows from Proposition \ref{prop:norm-transformation}. \qed



Theorem~\ref{thm:non-strong} presents convergence guarantees (that hold in expectation over the random selection rule) for Algorithms~\ref{al:rgbm} and~\ref{al:rgcd} for a non-strongly convex loss function $\ell$.

\smallskip

\begin{thm}\label{thm:non-strong}
Consider RGBM (Algorithm \ref{al:rgbm}) or equivalently {RtGCD} (Algorithm \ref{al:rgcd}) with either line-search step-size rule or constant step-size rule with $\rho={1}/{\sigma}$. If $\ell$ is a $\sigma$-smooth function and has a bounded level set, it holds for all $M\ge 0$ that
$$
\EE_{\xi_{M}}[\LC(\beta^{M})-\LC(\beta^{*})]\le\frac{1}{\frac{1}{\LC(\beta^{0})-\LC(\beta^{*})}+\frac{M}{2\sigma\D^2}} \le \frac{2\sigma\D^2}{M}\ .
$$
\end{thm}

Before presenting the proof of Theorem~\ref{thm:non-strong}, we present the 
following proposition, which is a generalization of the Cauchy-Schwarz inequality.

\smallskip

\begin{prop}\label{prop:CS}
For $a\in\Range(B^{T})$, it holds that
$$
\|a\|_{\FF}\DistFB(\beta,\hbeta)\ge\left\langle a,\beta-\hbeta\right\rangle\ .
$$
\end{prop}

\textbf{Proof.} Assume $a=B^{T}s$ and let $t=\argmin_{t\in Z(\hbeta)}\|\beta-t\|_{\FF^{*}}$
, then it holds that
\begin{equation*}
\begin{array}{lcl}
\|a\|_{\FF}\DistFB(\beta,\hbeta) & = &\|B^{T}s\|_{\FF}\|\beta-t\|_{\FF^{*}} \ge \left\langle B^{T}s,\beta-t\right\rangle =\left\langle s,B\beta-Bt\right\rangle\\ \\
 & =&\left\langle s,B\beta-B\hbeta\right\rangle =\left\langle B^{T}s,\beta-\hbeta\right\rangle =\left\langle a,\beta-\hbeta\right\rangle\ .  \qed
 \end{array}
\end{equation*}

\textbf{Proof of Theorem \ref{thm:non-strong}:}
Recall from \eqref{eq:decay} that for both step-size rules it holds that

\begin{equation}
\mathbb{E}_{m+1}[\LC(\beta^{m+1})]\le\LC(\beta^{m})-\frac{1}{2\sigma}\|\nabla\LC(\beta^{m})\|_{\FF}^{2} \ .\label{eq:middle-strong}
\end{equation}

Moreover, it follows from \eqref{eq:thm_eq_1-1} that $L(\beta^{m+1})\le L(\beta^m)$ (with probability one), thus for any iteration $m$, with probability one, we have $\beta^m\in\LS_0$. Noting that $\nabla\LC(\beta^{m})\in\Range(B^{T})$ and by using Proposition~\ref{prop:CS} we have

\begin{equation*}
    \begin{array}{lcl}
         \mathbb{\E}_{m+1}[\LC(\beta^{m+1})] & \le& \LC(\beta^{m})-\frac{\left\langle \nabla\LC(\beta^{m}),\beta^{m}-\beta^{*}\right\rangle ^{2}}{2\sigma\DistFB(\beta^{m},\beta^{*})^{2}}  
        \le  \LC(\beta^{m})-\frac{\left\langle \nabla\LC(\beta^{m}),\beta^{m}-\beta^{*}\right\rangle ^{2}}{2\sigma\D^{2}}\\ \\ & \le& \LC(\beta^{m})-\frac{(\LC(\beta^{m})-\LC(\beta^{*}))^{2}}{2\sigma\D^{2}}\ ,
    \end{array}
\end{equation*}

where the second inequality is because $\beta^m\in \LS_0$ (almost surely), and the third inequality follows from the convexity of $\LC$. Taking expectation with respect to $\xi_m$, we arrive
at
\begin{equation*}
    \begin{aligned}
\mathbb{\E}_{\xi_{m+1}}[\LC(\beta^{m+1})]&\le&\mathbb{\E}_{\xi_{m}}[\LC(\beta^{m})]-\frac{\E_{\xi_{m}}[(\LC(\beta^{m})-\LC(\beta^{*}))^{2}]}{2\sigma\D^{2}} \\ 
&\le& \mathbb{\E}_{\xi_{m}}[\LC(\beta^{m})]-\frac{\left(\E_{\xi_{m}}[\LC(\beta^{m})-\LC(\beta^{*})]\right)^{2}}{2\sigma\D^{2}}\ .
\end{aligned}
\end{equation*}

Now define $\delta_{m}:=\mathbb{\E}_{\xi_{m}}[\LC(\beta^{m})-\LC(\beta^{*})]$,
then we have $\delta_{m}\ge0$ and
\[
\delta_{m+1}\le\delta_{m}-\frac{\delta_{m}^{2}}{2\sigma\D^{2}}\ .
\]
Noticing that $\delta_{m+1}=\EE_{\xi_m}[\EE_{m+1}[\LC(\beta^{m+1})~|~\xi_m]\le\EE_{\xi_m}[\LC(\beta^m)]=\delta_m$, we have:
$$
\delta_{m+1}\le\delta_{m}-\frac{\delta_{m}\delta_{m+1}}{2\sigma\D^{2}} \ .
$$
Dividing both sides by $\delta_m \delta_{m+1}$, we arrive at
$$
\frac{1}{\delta_{m+1}} \ge \frac{1}{\delta_{m}} + \frac{1}{2\sigma\D^{2}}.
$$
Hence, we have that $$\frac{1}{\delta_{M}}\ge \frac{1}{\delta_{0}}+\frac{M}{2\sigma\D^2} \ ,$$ which completes the proof of the theorem.
\qed







\section{Numerical Experiments}\label{sec:numerical}
In this section, we present computational experiments discussing the performance of RGBM for solving classification and regression problems with tree stumps as weak-learners. Our code is publicly available at: \url{https://github.com/haihaolu/RGBM}.

{\bf Datasets:} The datasets we use in the numerical experiments were gathered from the LIBSVM library \cite{chang2011libsvm}. Table \ref{tab:stats} presents basic summary statistics of these datasets. For each dataset, we randomly choose $80\%$ as the training and the remaining as the testing dataset. In our experiments, we use the 
squared $\ell_2$ loss for the regression problem.
To be consistent with our theory (i.e., to have a strongly convex loss function), we use a regularized logistic loss with a small parameter $d=0.0001$ for the classification problems (see Section~\ref{sec:guarantees}).

\begin{table}[H]
\centering
\begin{tabular}{|c|c|c|c|}
\hline
dataset              & task           & \# samples      & \# features   \\ \hline
a9a               & classification & 32,561  & 123   \\ \hline
colon-cancer      & classification & 62     & 2,000  \\ \hline
rcv1              & classification & 20,242  & 47,236 \\ \hline
YearPrediction & regression     & 463,715 & 90    \\ \hline
\end{tabular}
\caption{\small{Basic statistics of the (real) datasets used in numerical experiments. The training/testing datasets are obtained by a 80\%/20\% (random) split on these sample sizes.}}
\label{tab:stats}
\end{table}

\begin{figure}[h!]
\centering
\resizebox{!}{0.48\textheight}{
\begin{tabular}{l c l c}
\multicolumn{4}{c}{ \sf  {~a9a dataset}} \medskip \\
\rotatebox{90}{\sf {~~~~~~~~~training loss}}&
\includegraphics[height=0.22\textheight,  trim =1.2cm 1.2cm 1.7cm 1.7cm, clip = true]{a9a_train_loss_time.png}&
\rotatebox{90}{\sf {~~~~~~~~~~~testing loss}}&
\includegraphics[height=0.22\textheight,  trim =1.2cm 1.2cm 1.7cm 1.7cm, clip = true]{a9a_test_loss_time.png} \medskip\medskip \\

\multicolumn{4}{c}{ \sf  {~colon-cancer dataset}} \medskip \\
\rotatebox{90}{\sf {~~~~~~~~~training loss}}&
\includegraphics[height=0.22\textheight,  trim =1.2cm 1.2cm 1.7cm 1.7cm, clip = true]{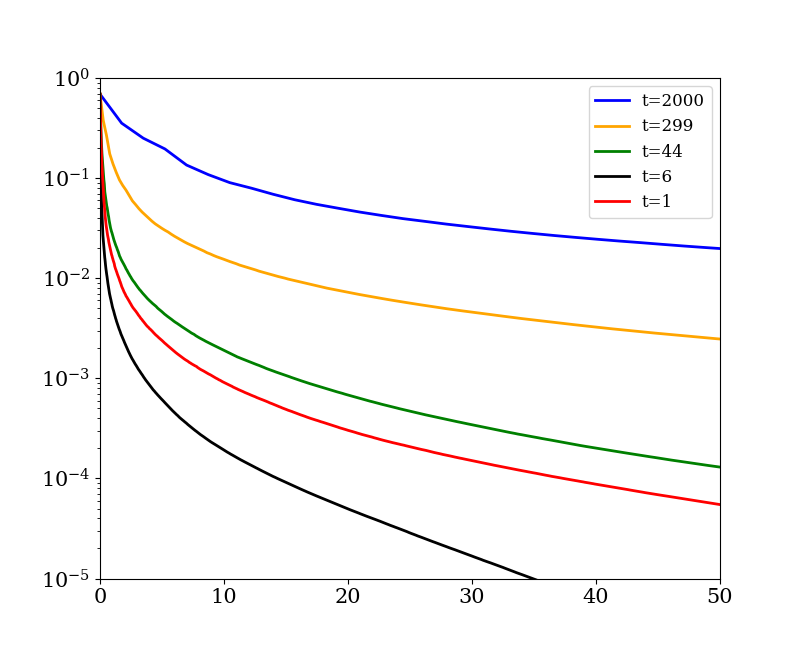}&
\rotatebox{90}{\sf {~~~~~~~~~~testing loss}}&
\includegraphics[height=0.22\textheight,  trim =1.2cm 1.2cm 1.7cm 1.7cm, clip = true]{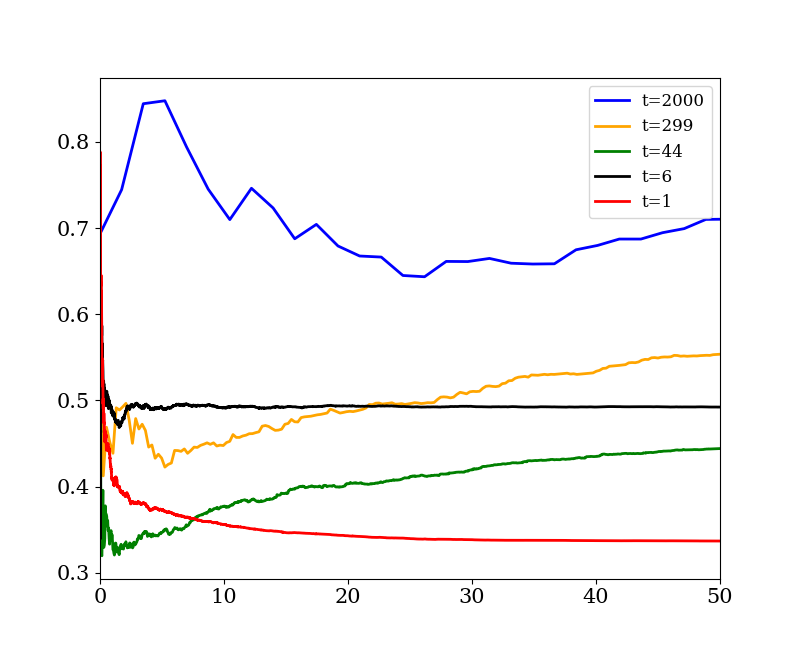} \medskip\medskip \\

\multicolumn{4}{c}{ \sf  {~rcv1 dataset}} \medskip \\
\rotatebox{90}{\sf {~~~~~~~~~training loss}}&
\includegraphics[height=0.22\textheight,  trim =1.2cm 1.2cm 1.3cm 1.7cm, clip = true]{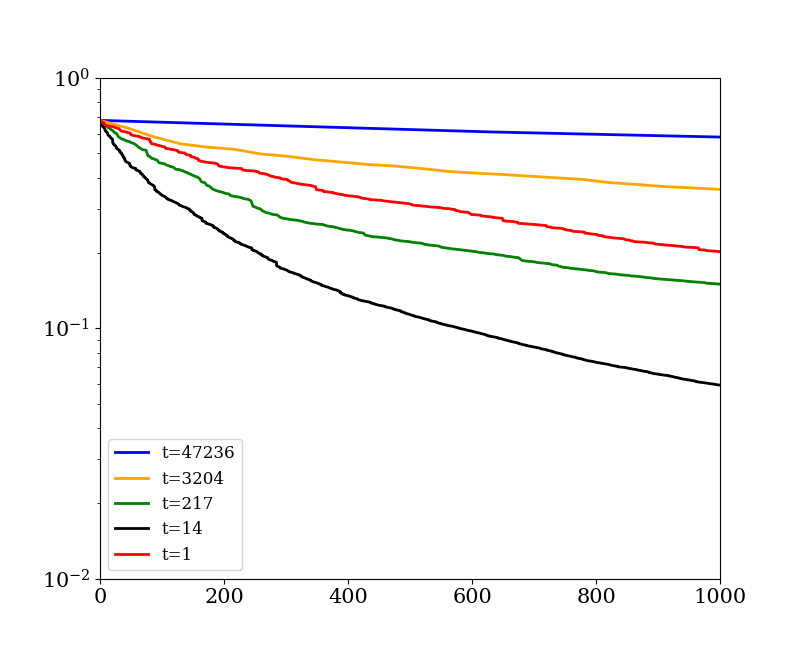}&
\rotatebox{90}{\sf {~~~~~~~~~~testing loss}}&
\includegraphics[height=0.22\textheight,  trim =1.2cm 1.2cm 1.3cm 1.7cm, clip = true]{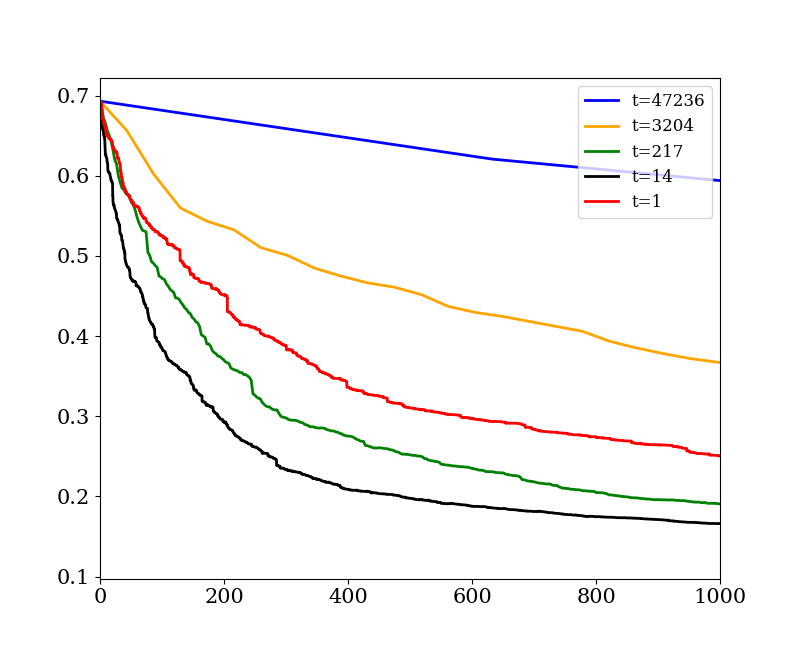} \medskip\medskip \\

\multicolumn{4}{c}{ \sf  {~YearPredictionMSD dataset}} \medskip \\
\rotatebox{90}{\sf {~~~~~~~~~training loss}}&
\includegraphics[height=0.22\textheight,  trim =1.2cm 1.2cm 1.5cm 1.7cm, clip = true]{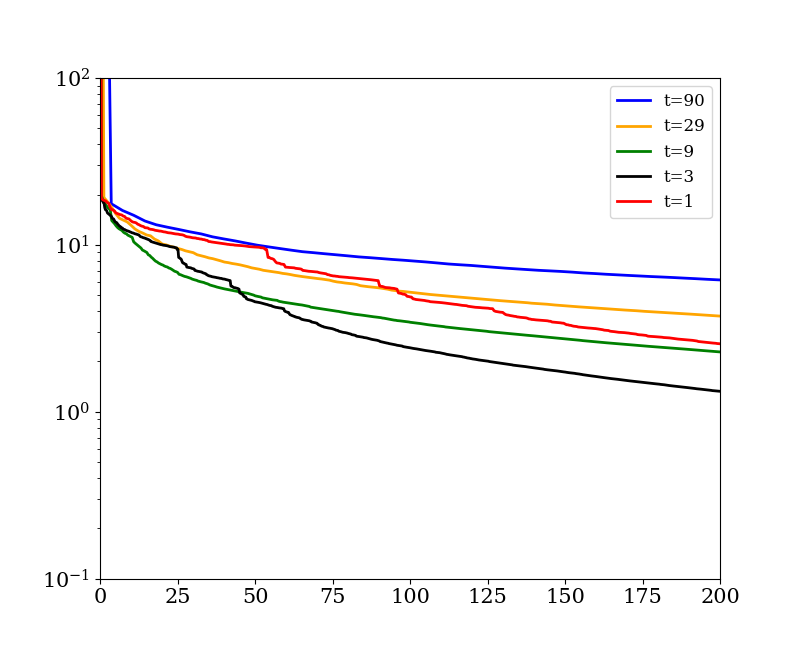}&
\rotatebox{90}{\sf {~~~~~~~~~~testing loss}}&
\includegraphics[height=0.22\textheight,  trim =1.2cm 1.2cm 1.5cm 1.7cm, clip = true]{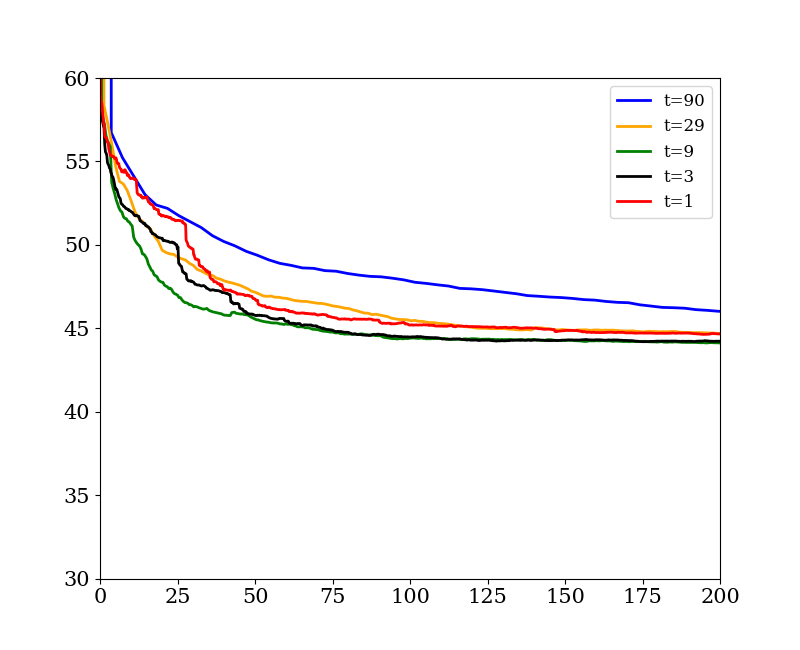} \medskip\medskip \\
  & \sf {Running Time} &    & \sf {Running Time} \\
\end{tabular}}
\vspace{-.5em}
\caption{{\small{Plots showing the training optimality gap (in $\log$ scale) and testing loss versus running time for four different datasets. We consider RGBM for different $t$ values (with the largest corresponding to GBM). The general observations are similar to that in Figure~\ref{fig:loss1}---we get significant computational savings by using a smaller value of $t$, without any loss in training/testing error. 
}}}\label{fig:loss_large}
\end{figure}


{\bf RGBM with Tree Stumps:} 
All algorithms consider tree stumps (see~\eqref{weak-learn-stmps}) as the weak-learners, as described in Section~\ref{sec:RGBM}. 
In our experiments (involving datasets with $n>10,000$), to reduce the computational cost, we 
decrease the number of candidate splits for each feature by considering $100$ quantiles instead 
of all $n$ quantiles (corresponding to $n$ samples). (We note that this simply reduces the number of weak learners considered, and our methodological framework applies.) This strategy is commonly used in implementations of GBM e.g, XGBoost~\cite{chen2016xgboost}. For each feature, we consider the candidate splitting points according to the percentiles of its empirical distribution, thus there are in total $100p$ weak-learners. All the tree stumps that perform a split on one feature is considered as a group---thereby, resulting in $p$ groups. In RGBM, we randomly choose $t$ out of $p$ features and consider the $100t$ features as the set $J$, among which we pick the best weak-learner to perform an update. The values of $t$ are chosen on a geometrically spaced grid from $1$ to $p$ with five values for each dataset. In particular, the case $t=p$ corresponds to traditional GBM.

{\bf Performance Measures:}
Figure \ref{fig:loss_large} shows the performance of RGBM with different $t$ values. The x-axis is the running time (in seconds).
All computations were carried out on MIT Sloan's Engaging Cluster on an Intel Xeon 2.30GHz machine (one CPU) with 10GB of RAM memory.   
The y-axis denotes the quality of solution (or the data-fidelity) obtained, i.e., the objective value, for both the training and testing datasets.

{\bf Comparisons:}
For all datasets, RGBM with a medium $t$ value leads to a model with the smallest training error with the same running time. This demonstrates the (often significant) computational gains possible by using RGBM. 
For datasets with $n \gg p$, the profile of the testing loss is similar to that of the training loss. The colon-cancer dataset is a high-dimensional problem with $p\gg n$; and its training/testing profile is somewhat different from the other datasets---here, the model with the best test error corresponds to a moderately large training error, and models with the smallest training error lead to poor generalization. In many examples, we observe that a choice of $t$ in the interior of its range of possible values, leads to a model with best test performance. For many examples (e.g, the third and fourth rows), we see that the testing error profiles saturate near the minimum even if the training error continues to decrease. The last two observations empirically suggest that the randomization scheme within RGBM potentially imparts additional regularization
resulting in good generalization performance. 

\section{Discussion}
\textcolor{black}{In this paper we present a greedy coordinate descent perspective of the popular GBM algorithm, where the coordinates correspond to weak learners and the collection of weak learners/coordinates can be exponentially large. 
We introduce and formally analyze RGBM, a randomized variant of popular GBM. 
RGBM can be thought as a random-then-greedy coordinate descent procedure where, we randomly select a subset of weak-learners and then choose the best weak learner from these candidates by a greedy mechanism. This presents a formal algorithmic justification of common heuristics used within the popular GBM implementations (e.g., XGBoost). From an optimization perspective, RGBM can be interpreted as a natural bridge between greedy coordinate descent on one end and randomized coordinate descent on the other. Our random-then-greedy coordinate descent procedure can be used as a standalone algorithm; and can be potentially employed in machine learning contexts where coordinate descent is a popular choice~\cite{wright2015coordinate}. On a related note, recent developments in large scale coordinate descent---such as the works of~\cite{scherrer2012feature} and~\cite{stich2017approximate} may be used to improve upon our proposed coordinate descent procedure (and in particular, RGBM).
We derive new computational guarantees for RGBM based on a coordinate descent interpretation. The guarantees depend upon a quantity that we call MCA (Minimum Cosine Angle) relating to the density of the weak-learners or basis elements in the prediction space. 
The MCA quantity seems to bear some similarities with Cheung-Cucker condition number~\cite{cheung2001new,ramdas2016towards} used to analyze computational guarantees of solving feasibility problems in a linear system. A precise connection between these quantities does not seem to be straightforward; and a detailed investigation of their links is an interesting direction for future research.\\
The focus of our paper is on the algorithmic properties of RGBM---in terms of minimizing the empirical loss function as opposed to the population risk. 
Boosting is empirically known to lead to excellent out-of-sample properties by virtue of its implicit regularization properties~\cite{ESLBook,telgarsky2012primal,zhang2005boosting} that are imparted by the algorithm. As our numerical experiments suggest, RGBM appears to have superior generalization properties by virtue of its random-then-greedy selection rule as opposed to a pure greedy method (as in GBM). 
A formal explanation of the generalization ability of RGBM is not addressed in this work, and is an important topic of future research.} 

\vspace{-.6em}

\subsection*{Acknowledgements}
The authors would like to thank Robert Freund for helpful discussions and feedback. The authors would also like to thank the Associate Editor and the Reviewers for their helpful feedback that helped us improve the paper. The research is supported partially by ONR-N000141512342, ONR-N000141812298 (YIP) and NSF-IIS1718258.

\vspace{-1em}

\bibliographystyle{amsplain}

{\small{\bibliography{LF-papers}}}

\end{document}